\pdfoutput=1
\documentclass[11pt]{article}
\usepackage[final]{acl}

\usepackage{times}
\usepackage{url}
\usepackage{latexsym}
\usepackage[T1]{fontenc}
\usepackage[utf8]{inputenc}
\usepackage{xspace,microtype}
\usepackage{array}  %
\usepackage{graphicx}
\usepackage{tcolorbox}
\usepackage{arydshln}
\usepackage{comment}
\usepackage{subcaption}
\usepackage{dsfont}
\usepackage{amsmath}
\usepackage{amsfonts}
\usepackage{amssymb}
\usepackage{amsthm}
\usepackage{mathtools}
\usepackage{bm}
\usepackage[linesnumbered,ruled,vlined]{algorithm2e}
\SetAlgoCaptionLayout{small}
\usepackage{listings}
\usepackage{CJKutf8}

\usepackage{enumitem}

\usepackage{tikz}
\usetikzlibrary{calc}
\usepackage[hang,flushmargin]{footmisc}

\usepackage[table]{xcolor}
\usepackage{colortbl}
\usepackage{collcell}
\usepackage{xcolor}
\usepackage{booktabs}
\usepackage{multirow}
\definecolor{cK}{RGB}{221,221,221}
\definecolor{cR}{RGB}{255,204,204}
\definecolor{cG}{RGB}{204,255,204}
\definecolor{cB}{RGB}{204,204,255}
\definecolor{navyblue}{rgb}{0.0,0.0,0.5}
\definecolor{darkred}{rgb}{0.65, 0, 0}
\definecolor{darkgreen}{rgb}{0, 0.5, 0}
\definecolor{lB}{rgb}{0,0.4,0.9}
\definecolor{verylightblue}{RGB}{230,240,255}
\definecolor{lightblue}{RGB}{200,220,255}
\definecolor{mediumblue}{RGB}{150,190,255}
\definecolor{sig10}{RGB}{255,245,220} 
\definecolor{sig05}{RGB}{255,225,180}
\definecolor{sig01}{RGB}{255,180,100}

\newcommand{\Figure}[1]{Figure~\ref{#1}}
\newcommand{\Table}[1]{Table~\ref{#1}}
\newcommand{\Section}[1]{§\ref{#1}}

\newcommand{\formatrl}{\textls*[-20]{\textsc{Format}}\kern-0.05em\textsc{RL}}
\newcommand{\bleu}{\textls*[-12]{\textsc{Bleu}}\xspace}

\usepackage{soul}
\sethlcolor{yellow}

\definecolor{myblue}{HTML}{2c7fb8}

\newcommand{\argmin}{\mathop{\rm arg~min}\limits}
\usepackage{hyperref}

\title{Structured Document Translation via\\Format Reinforcement Learning}

\author{
 \textbf{Haiyue Song\textsuperscript{1}},\;
 \textbf{Johannes Eschbach-Dymanus\textsuperscript{2}},\;
 \textbf{Hour Kaing\textsuperscript{1}},\;
 \textbf{Sumire Honda\textsuperscript{2}},\\
 \textbf{Hideki Tanaka\textsuperscript{1}},\;
 \textbf{Bianka Buschbeck\textsuperscript{2}},\;
 \textbf{Masao Utiyama\textsuperscript{1}}
\\
 \textsuperscript{1}National Institute of Information and Communications Technology, Japan\;
 \textsuperscript{2}SAP, Germany
 \\
 \texttt{\{haiyue.song,hour\_kaing,hideki.tanaka,mutiyama\}@nict.go.jp}\\
 \texttt{\{johannes.eschbach-dymanus,sumire.honda,bianka.buschbeck\}@sap.com}
}

\begin{document}
\maketitle
\begin{abstract}
Recent works on structured text translation remain limited to the sentence level, as they struggle to effectively handle the complex document-level XML or HTML structures.
To address this, we propose \textbf{Format Reinforcement Learning (\formatrl{})}, which employs Group Relative Policy Optimization on top of a supervised fine-tuning model to directly optimize novel structure-aware rewards: 1) TreeSim, which measures structural similarity between predicted and reference XML trees and 2) Node-chrF, which measures translation quality at the level of XML nodes.
Additionally, we apply StrucAUC, a fine-grained metric distinguishing between minor errors and major structural failures.
Experiments on the SAP software-documentation benchmark demonstrate improvements across six metrics and an analysis further shows how different reward functions contribute to improvements in both structural and translation quality.

\end{abstract}

\section{Introduction}
\begin{figure}[t] 
  \centering
  \includegraphics[width=0.99\linewidth]{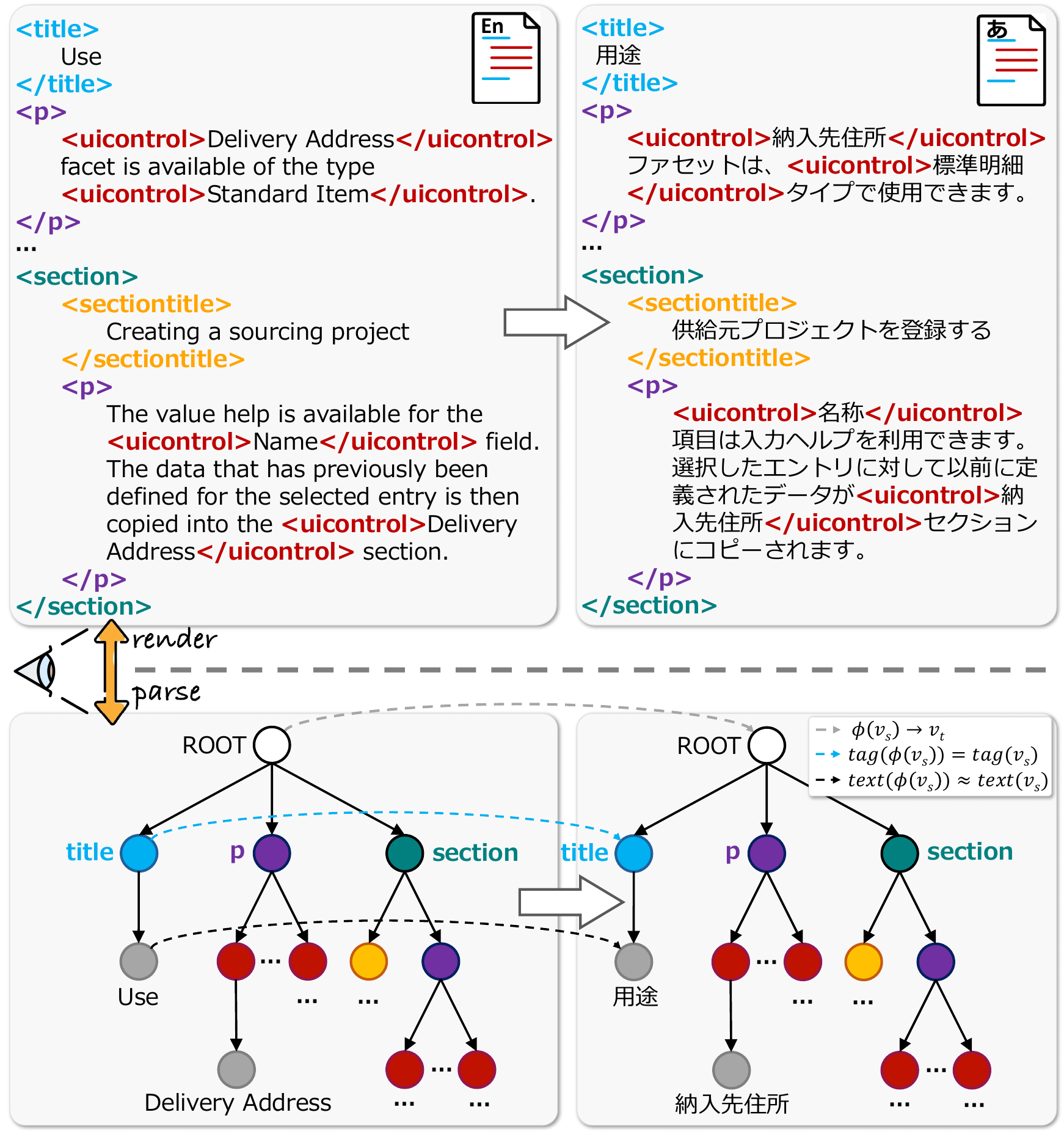}
  \caption{A structured document translation example (English$\rightarrow$Japanese), with markup highlighted in color. The lower part shows the translation of XML tree structure with node $\phi(\mathord{\cdot})$, $\text{tag}(\mathord{\cdot})$, and $\text{text}(\mathord{\cdot})$ mappings.}
  \label{fig:task}
\end{figure}

Translating structured documents such as software manuals is essential for product localization. As shown in~\Figure{fig:task}, they carry markup that defines layout and interactive elements, making structural fidelity as important as content translation quality. 

Until the advent of large language models (LLMs), the most prevalent approach for translation with markup was the detag-and-project pipeline~\cite{joanis-etal-2013-transferring,muller-2017-treatment,zenkel-etal-2021-automatic-bilingual}. This pipeline usually leverages a machine translation (MT) system to translate plain text (with tags removed) and a separate word aligner to reinsert the tags into the translated text. Although straightforward, it is prone to error propagation from individual MT and alignment modules.

LLMs have emerged as a promising end-to-end solution for markup translation~\cite{dabre-etal-2023-study,dabre-etal-2024-effective}. 
Few-shot prompting is a convenient way to enable LLMs to learn markup transfer patterns with only a few examples~\cite{brown2020language,lewis2020retrieval,dabre-etal-2023-study}, and fine-tuning provides more robust domain adaptation capabilities thus better performance~\cite{dabre-etal-2024-effective}.
However, the training objective of supervised fine-tuning is to optimize token-level likelihood, leaving markup accuracy largely unaddressed. Therefore, it is difficult for them to handle complex structured documents such as the one shown in~\Figure{fig:task}. 

In this study, we address these limitations by proposing Format Reinforcement Learning (\formatrl{}), which moves from the token-level likelihood optimization to directly optimizing structure-aware objectives. It first fine-tunes an LLM for basic document translation capability, then applies Group Relative Policy Optimization (GRPO) with two novel structure-aware rewards: TreeSim for measuring XML tree structural similarity via edit distance, and Node-chrF for node-level translation quality assessment. The main contributions of this paper are summarized below:
\begin{itemize}[topsep=2pt, partopsep=2pt, itemsep=2pt, parsep=2pt]
    \item We propose Format Reinforcement Learning (\textbf{\formatrl{}}) for structured document translation. It utilizes Group Relative Policy Optimization (GRPO)~\cite{2402.03300} to optimize structural fidelity through novel structure-aware rewards \textbf{TreeSim} and \textbf{Node-chrF}. We also investigate a range of additional rewards to reinforce structural fidelity and translation quality.
    \item We use a new metric, Structure-Aware Area Under Curve (\textbf{StrucAUC}), which distinguishes between minor errors and major failures, then robustly combines both translation and structural quality into a single score.
    \item Our experimental results demonstrate significant improvements on the software documentation dataset~\cite{sap-software-documentation-data} across four translation directions, with \formatrl{} achieving average gains of $3.69$ XML-Match, $2.16$ XML-\bleu, $0.22$ Content-\bleu, and $0.93$ StrucAUC scores compared to a strong supervised fine-tuning baseline.
\end{itemize}

\section{Related Work}
This study focuses on \Section{sec:related-struct} structured text translation and \Section{sec:related-rl} reinforcement learning.

\subsection{Structured Text Translation}
\label{sec:related-struct}

The traditional detag-and-project approaches rely on separate modules for translation and markup handling~\cite{du-etal-2010-tmx,joanis-etal-2013-transferring,muller-2017-treatment,hanneman-dinu-2020-markup,zenkel-etal-2021-automatic-bilingual,ryu-etal-2022-data,steffen-van-genabith-2021-transins, zenkel-etal-2021-automatic-bilingual}. 
However, these methods suffer from error propagation across modules, and the MT system translates at the sentence level without using document-level context.

Recent end-to-end approaches for structured text translation have become possible with LLMs such as BLOOM~\cite{scao2022bloom}, ChatGPT~\cite{brown2020language,openai2023gpt4}, and Llama 3~\cite{dubey2024llama3}, 
owing to their strong in-context learning and generalization capabilities.
Previous studies using few-shot prompting~\cite{dabre-etal-2023-study} or fine-tuning on a small dataset~\cite{dabre-etal-2024-effective} work well for sentence translation with markup. However, they struggle to handle complex structures such as those found in XML documents.

\subsection{Reinforcement Learning}
\label{sec:related-rl}
From the perspective of RL, generation is a sequence of actions to maximize the reward. 
Proximal Policy Optimization (PPO)~\cite{1707.06347} is the RL algorithm used in ChatGPT~\cite{openai2023gpt4}, whereas GRPO~\cite{2402.03300} used in DeepSeek-R1~\cite{2501.12948} further simplifies PPO by removing the separate value network.
RL has been applied to tasks such as code generation~\cite{dou-etal-2024-stepcoder}, JSON generation~\cite{2502.18878} and format instruction following~\cite{2412.09173}. 
To our knowledge, we are the first to apply RL algorithms to the structured document translation task, whose challenge lies in designing rewards to guide generation of exactly the same structure as that in the source document while maintaining the high translation quality.

\begin{figure*}[t] 
  \centering
  \begin{tikzpicture}
    \node[anchor=south west,inner sep=0] (image) at (0,0) {\includegraphics[width=0.93\linewidth]{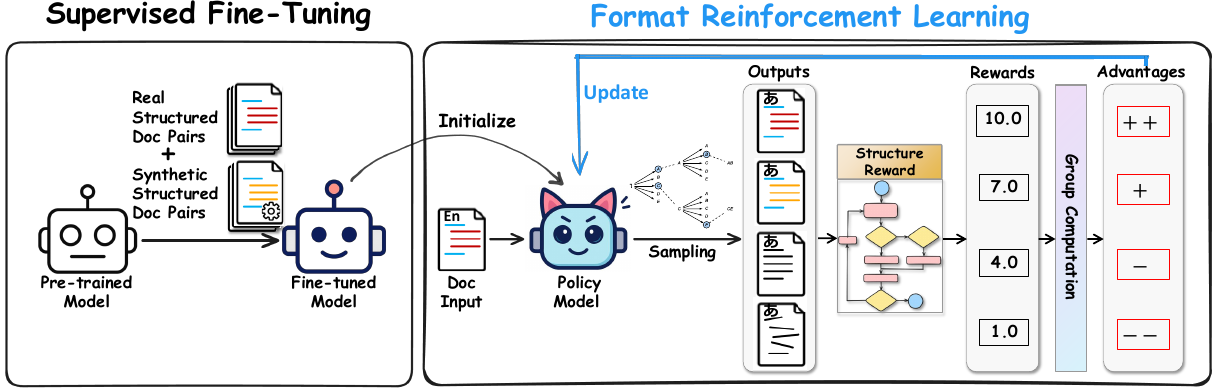}};
    \node[anchor=center, minimum width=0.5cm, minimum height=0.270cm, 
          fill=blue, fill opacity=0.00, text opacity=1] 
          (sft-area) at ($(image.south west) + (0.013\linewidth, 0.286\linewidth)$) 
          {\hyperref[sec:sft]{\textcolor{blue}{\S\ref{sec:sft}}}};
    \node[anchor=center, minimum width=0.5cm, minimum height=0.270cm,
          fill=blue, fill opacity=0.00, text opacity=1]
          (rl-area) at ($(image.south west) + (0.428\linewidth, 0.285\linewidth)$)
          {\hyperref[sec:rl]{\textcolor{blue}{\S\ref{sec:rl}}}};
  \end{tikzpicture}
  \caption{Our \formatrl{} pipeline consists of two stages. First, we fine-tune a pre-trained LLM (e.g., Llama-3.1-8B-Instruct) using real and synthetic structured document pairs. Second, we reinforce the format handling ability through GRPO with our proposed format reward functions.} %
  \label{fig:pipeline}
\end{figure*}

\section{Method}
\label{sec:method}
Our pipeline is shown in~\Figure{fig:pipeline}. 
We first define the task in~\Section{sec:task}, then describe the supervised fine-tuning (SFT) phase in~\Section{sec:sft}, and finally present the core reinforcement learning phase in~\Section{sec:rl}.

\subsection{Task Definition}
\label{sec:task}
This work addresses the task of translating a structured document \(D_s\) in the source language into its counterpart \(D_t\) in the target language.
A structured document \(D\) can be viewed as an XML tree \(D = (V_D, E_D)\), where \(V_D\) denotes the set of nodes and \(E_D\) the set of parent–child edges. 
Each node is associated with a tag symbol $\textit{tag}(v)$ (e.g., \texttt{<p>}) and may contain textual segments $\textit{text}(v)$.

The translation model \(\pi_\theta\) is a conditional probability distribution defined as follows:
\[
\pi_\theta: \mathcal{D}_s \times \mathcal{D}_t \rightarrow [0,1] \subset \mathbb{R}, \quad \pi_\theta(D_t \mid D_s)
\]
where \(\pi_\theta(D_t \mid D_s)\) denotes the probability of generating the target document \(D_t\) given the source document \(D_s\), and $\mathcal{D}_s$ and $\mathcal{D}_t$ are the spaces of all possible structured documents in the source and target languages.
The predicted translation \(\hat{D}_t\) is typically obtained by maximizing this probability:
\[
\hat{D}_t = \arg\max_{D_t \in \mathcal{D}_t} \pi_\theta(D_t \mid D_s)
\] 
We assume that the predicted document \(\hat{D}_t\) satisfies the following two conditions we target:
\begin{enumerate}[topsep=2pt, partopsep=2pt, itemsep=2pt, parsep=2pt]
\item \textbf{Structural Identity}: \(\hat{D}_t\) is isomorphic to the source tree \(D_s\). Formally, there exists a bijection \(\phi: V_{D_s} \rightarrow V_{\hat{D}_t}\) such that:
 \begin{itemize}[topsep=2pt, partopsep=2pt, itemsep=2pt, parsep=2pt]
    \item For any edge \((u, v) \in E_{D_s}\), we have \((\phi(u), \phi(v)) \in E_{\hat{D}_t}\). 
    \item For any internal node \(v \in V_{D_s}\), the corresponding target node shares the same tag symbol: \(\textit{tag}(\phi(v)) = \textit{tag}(v)\).
 \end{itemize}

\item \textbf{Translation Correspondence}: For each source node \(v \in V_{D_s}\) and its corresponding target node \(\phi(v)\) their textual contents \(\textit{text}(v)\) and \(\textit{text}(\phi(v))\) are mutual translations.

\end{enumerate}

To examine the extent to which both conditions are satisfied, in practice, we measure the translation quality between the predicted tree \(\hat{D}_t\) and a reference document \(D_t^\star\) using well-established metrics such as \bleu~\cite{papineni2002bleu} and COMET~\cite{rei2020comet,rei-etal-2022-comet}.

\subsection{Phase I: Supervised Fine-Tuning}
\label{sec:sft}
We fine-tune a pre-trained LLM on parallel structured documents. To address the data scarcity problem, we synthesize training data by injecting XML markup into parallel plain-text documents.

\paragraph{Data Synthesis.}
Given a parallel corpus of plain documents $\{(d_s^i, d_t^i)\}_{i=1}^N$, we use GPT-4o to generate structured documents $\{(D_s^i, D_t^i)\}_{i=1}^M$ in which:
\begin{itemize}[topsep=2pt, partopsep=2pt, itemsep=2pt, parsep=2pt]
    \item Both $D_s^i$ and $D_t^i$ have the same structure;
    \item The original parallel texts are preserved.
\end{itemize}

We ensure structural identity through validation: for each generated pair $(D_s, D_t)$, we verify their XML trees are isomorphic. Invalid pairs are regenerated until success or hitting the retry limit.

\subsection{Phase II: Format Reinforcement}
\label{sec:rl}
Initialized from the SFT checkpoint, we use our designed rewards to optimize the translation model (policy model as termed in GRPO) to generate structurally correct and high-quality translations.

\subsubsection{Reward Functions}
The policy model learns from good samples generated by itself during training, where the reward function defines what is good. During GRPO training, a reward function $r(\hat{D}_{t,i}, D_t^\star)$ compares each sampled output $\hat{D}_{t,i} \sim \pi_\theta(\cdot|D_s)$ with the reference document $D_t^\star$, and indicates how good each output is.
To reinforce structure-aware similarity, we propose two rewards: TreeSim and Node-chrF.
\paragraph{TreeSim} measures structural similarity between the predicted and reference XML trees.
It first parses both documents as XML fragments wrapped in a dummy root.
The similarity is computed using the Zhang-Shasha tree edit distance~\citep{zss}, which counts the minimum number of node insertions, deletions, or relabelings needed to transform one tree into another.
To obtain a normalized similarity score, we use:
$$\text{TreeSim}(\hat{D}_{t,i}, D_t^\star) = 1 - \frac{\text{EditDist}(\hat{D}_{t,i}, D_t^\star)}{\max(|\hat{D}_{t,i}|, |D_t^\star|)}$$
where $\text{EditDist}$ is the tree edit distance and $|D|$ denotes the number of nodes in tree $D$ excluding the dummy root.
This normalization ensures that the score remains in $[0, 1]$, with 1 indicating identical structures and 0 maximum dissimilarity. Specifically, we assign a penalty score of $-0.1$ for invalid XML that cannot be parsed. 

\paragraph{Node-chrF} measures translation quality at the level of individual XML nodes.
The algorithm performs a parallel depth-first traversal of the predicted and reference XML trees, pairing nodes at corresponding positions. When the sizes differ, the algorithm extends the shorter traversal list to match the longer one by adding empty placeholders.
For each node pair $(v_{\text{pred}}, v_{\text{ref}})$, the metric computes:
\begin{itemize}[topsep=1pt,itemsep=1pt,parsep=1pt]
    \item A score of $0$ if the nodes have mismatched tags (e.g., \texttt{<p>} vs \texttt{<h1>}) or are unpaired (e.g., one subtree has more nodes than the other)
    \item The chrF score~\citep{popovic-2015-chrf} of their textual content (excluding child nodes) if tags match
    \item Skip node pairs that contain only whitespace
\end{itemize}
The final score is the average of all node pairs:
\vspace{-0.5em}
\begin{align*}
\text{N}&\text{ode-chrF} = \\
&\frac{1}{|\mathcal{P}|} \sum_{\mathrlap{\hspace{-1em}(v_{\text{pred}}, v_{\text{ref}}) \in \mathcal{P}}} \mathds{1}_{\text{match}}(v_{\text{pred}}, v_{\text{ref}})\cdot\text{chrF}(v_{\text{pred}}, v_{\text{ref}})
\end{align*}
, where $\mathcal{P}$ is the set of all node pairs in the traversal and $\mathds{1}$ is the indicator function for tag matching. For aligned trees, this metric focuses on translation quality. If the translation contains structural mistakes, however, nodes become misaligned, and the reward degrades substantially.

In practice, we scale each reward to $|r| \in [0, 10]$ for numerical stability. 
We also investigate the use of other metrics (\Section{sec:ablation-reward}) as rewards and explore combining two rewards (by summing the scores).

\subsubsection{Optimization}
After calculating reward scores for a group of samples, we encourage the model to generate similar high-scoring outputs.
In GRPO, we calculate the relative performance comparisons within the group, called advantages, which is then used to update the document translation policy model $\pi_\theta$.

Formally, the optimization process works as follows: for each source document $D_s$, we generate $K$ candidate translations $\{\hat{D}_{t,i}\}_{i=1}^K$ from the current policy $\pi_\theta$. Instead of requiring absolute quality assessments, GRPO computes advantages by comparing each generation's reward against the group mean, effectively learning which translations are better than average within the same context.
Since we perform a single gradient update per exploration stage when computing gradients, we can remove the min and clip operation. This yields the following objective:
\begin{subequations}
\label{eq:grpo_loss_simplified}
\begin{align}
\mathcal{L}_{\text{GRPO}} = 
&-\mathbb{E}_{D_s \sim \mathcal{D}, \{\hat{D}_{t,i}\}_{i=1}^K \sim \pi_\theta(\cdot|D_s)} \nonumber \\[-2pt]
&\quad \left[ {\frac{1}{K} \sum_{i=1}^{K} \hat{A}_i \log \pi_\theta(\hat{D}_{t,i}|D_s)} \right] \label{eq:grpo_policy_gradient} \\
&+ \beta \cdot D_{KL}(\pi_\theta || \pi_{\text{SFT}}) \label{eq:grpo_kl_penalty}
\end{align}
\end{subequations}

The first term~\eqref{eq:grpo_policy_gradient} encourages the model to increase the likelihood of generations with positive advantages and to decrease the likelihood of those with negative advantages, with $\hat{A}_i$ computed as:

{\small
\begin{equation*}
\begin{aligned}
\hat{A}_i &= \frac{r(\hat{D}_{t,i}, D_t^\star) - \bar{r}}{\sigma_r} \\
\bar{r} &= \frac{1}{K}\sum_{j=1}^{K} r(\hat{D}_{t,j}, D_t^\star) \\
\sigma_r &= \sqrt{\frac{1}{K}\sum_{j=1}^{K} (r(\hat{D}_{t,j}, D_t^\star) - \bar{r})^2}
\end{aligned}
\end{equation*}
}
The second term~\eqref{eq:grpo_kl_penalty} is a Kullback-Leibler divergence regularizer that prevents the optimized policy $\pi_\theta$ from deviating too far from the supervised fine-tuned model $\pi_{\text{SFT}}$ with $\beta$ controlling its strength, thereby avoiding catastrophic forgetting.

\section{Evaluation Metrics: StrucAUC}
\label{sec:strucauc}
Previous studies on structured data translation apply the XML-\bleu metric~\cite{salesforce-xml-bleu} as a combined score for both translation quality and structural fidelity. However, it results in a zero score on the document-level even with minor structural mismatch. 
To provide a more fine-grained evaluation, we propose StrucAUC that distinguishes between minor errors and major structural failures. In detail, it provides a translation quality evaluation by interpolating between two scores, Node-chrF and Optimal Node-chrF, to measure quality with different levels of error tolerance. 

Optimal Node-chrF provides a way to measure Node-chrF for two documents with slightly different structures by node alignment.
It represents each node by its entire subtree (including tags and descendants) and computes a cost matrix $C$, where $C_{ij}$ is the chrF distance between the $i$-th node's subtree in $\hat{D}_t$ and the $j$-th node's subtree in the $D^\star_t$.
Using the Hungarian algorithm~\citep{kuhn1955hungarian}, it solves the linear sum assignment problem to find the optimal one-to-one mapping $\mathcal{M}^*$ that minimizes the total distance:
$$\mathcal{M}^* = \argmin_{\mathcal{M}} \sum_{(v_{\text{pred}}, v_{\text{ref}}) \in \mathcal{M}} C_{v_{\text{pred}}, v_{\text{ref}}}$$
The final score evaluates chrF on node-level textual content (excluding children) under this optimal alignment:
Nodes that cannot be matched are scored 0, ensuring all nodes are accounted for.

StrucAUC then integrates structural tolerance through tree edit distance~\citep{zss} into Optimal Node-chrF. For each document, we calculate the minimum number of edits required to transform the predicted tree into its optimally aligned version (as determined by $\mathcal{M}^*$), with tag mismatches counting as 0.5 edits. The metric then computes a curve at the corpus level: at each edit threshold $k \in \{0, 0.5, 1, ..., K\}$, documents requiring at most $k$ edits contribute their Optimal Node-chrF score, while others contribute their regular Node-chrF score. The area under this curve from 0 to $K$ yields StrucAUC@$K$, providing a smooth degradation from perfect structural alignment to increasing structural deviations. This makes StrucAUC robust for document-level evaluation: minor structural errors (e.g., a misplaced formatting tag) result in proportional score reductions rather than complete failure, while still rewarding structural fidelity. In our experiments, we report StrucAUC@5, allowing up to 5 structural edits before considering a document structurally misaligned. We provide the pseudo code in Appendix~\ref{app:strucauc}.

\section{Experimental Settings}
This section describes our dataset, evaluation metrics, and implementation details of our method.

\subsection{Dataset}
We use the SAP software documentation dataset \cite{sap-software-documentation-data} that contains parallel structured documents for language pairs including Japanese--English and Chinese--English translated by professional translators. Each language pair consists of $190$ document pairs for testing, and an additional $195$ document pairs, of which we use $100$ for training and $95$ for development. Each source--target document pair contains the same number of lines with a one-to-one, linear alignment, reflecting the property of this task that the page layout in different languages should be identical.

\paragraph{Statistics} Documents in this dataset exhibit substantial structural variety. After converting documents into XML trees, each tree has an average depth of $7.11$~$\pm$~$1.51$ and contains $27.36$~$\pm$~$25.28$ nodes, with a median of $18$ nodes per document, and an average of $14.62$ text segments per document. Overall, it covers $58$ unique XML tags.

\begin{figure}[h] 
  \centering
  \includegraphics[width=0.99\linewidth]{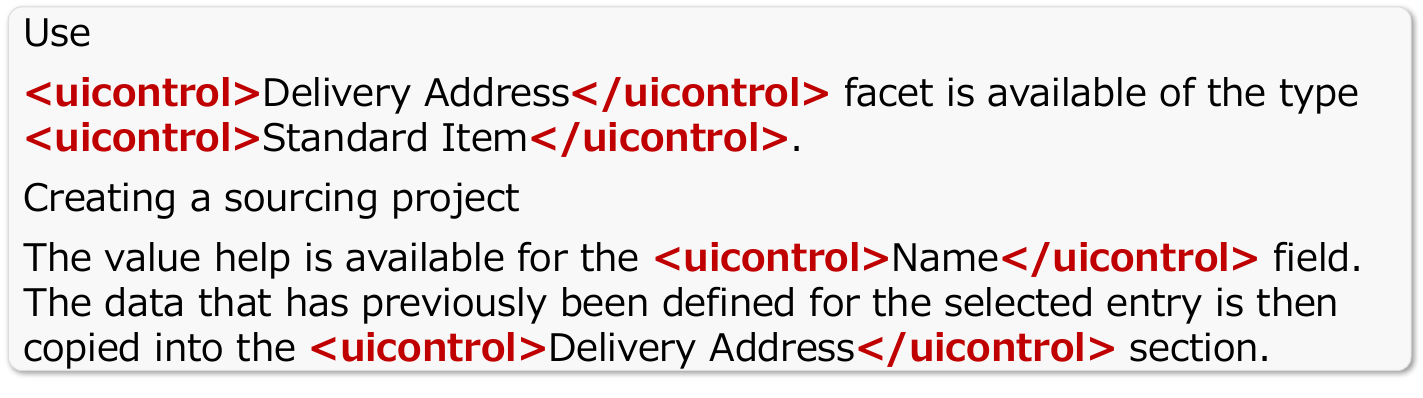}
  \caption{Inline markup version of the example in~Fig.\ref{fig:task}.}
  \label{fig:inline-example}
\end{figure}

\paragraph{Inline Markup Setting} The dataset also provides a simplified version with only sentence-internal markup, as shown in~\Figure{fig:inline-example}. We name it the \textit{inline markup setup}. 
Different from the structured setup which preserves non-translatable nodes (e.g., \texttt{<source>In-App Help</source>} and metadata), the inline setup keeps only translatable spans with inline tags. Consequently the reference texts also differ between the two setups.
We construct the data for both setups with the official SAP XSLT preprocessing scripts.

\subsection{Evaluation}
We apply six evaluation metrics, including four from previous studies: Content-\bleu, XML-Validity, XML-Match, XML-\bleu, and two proposed metrics: Content-COMET and StrucAUC. We classify them into three categories: 1) Translation: these mainly measure translation quality, 2) Structure: the ones measure structure fidelity, and 3) Combined: the ones measure both.
\paragraph{Translation} Content-\bleu is the \bleu for a document with all XML markup removed. We employ the SacreBLEU tool~\cite{post-2018-call} with language-specific tokenizers.\footnote{e.g., signature for Japanese: "nrefs:1|case:lc|eff:no|tok:ja-mecab-0.996-IPA|smooth:exp|version:2.5.1"}
\textbf{Content-COMET} is based on the neural MT metric COMET-22~\cite{rei-etal-2022-comet}. The metric is applied to the document texts without XML markup, as COMET-22 was not trained on structured documents and therefore cannot be directly applied to such data.
\paragraph{Structure} XML-Validity returns a binary score of one or zero whether the output $D_t$ passed XML parsing. XML-Match is also binary indicating whether the XML trees of output $D_t$ and reference $D_t^\star$ are exactly the same.
\paragraph{Combined} The XML-\bleu metric~\cite{salesforce-xml-bleu} is a combined score for both translation quality and structural fidelity. First, both translated and reference documents are split into text segments at XML tag boundaries. If a translation's XML-Match is true, the segments are paired for \bleu computation. Otherwise, the reference's segments are paired with empty strings, thereby penalizing structural errors. The metric is then computed on the corpus level across all segments of all documents. \textbf{StrucAUC} is also a combined metric as described in \Section{sec:strucauc}.

Additionally, we report empirical $p$-values from statistical significance testing using bootstrap resampling with $1,000$ trials.

\subsection{Implementations}
We report implementation details and hyperparameters selected based on our preliminary experiments.

\subsubsection{Prompting Baseline}
For the few-shot prompting baseline~\cite{dabre-etal-2023-study}, we evaluate pre-trained language models (GPT-4o and meta-llama/Llama-3.1-8B-Instruct) using in-context learning with $k \in \{0, 1, 2, 3, 4, 5\}$ full document pairs as exemplars, and report the $k$-shot setting with the highest XML-\bleu. We use greedy decoding for deterministic outputs and set the maximum generation length to 2,000 tokens which is sufficient to accommodate long document translations. 

\subsubsection{Supervised Fine-Tuning}
This section describes the LLM fine-tuning method as in~\citet{dabre-etal-2024-effective} on our document-level data. We use Llama-3.1-8B-Instruct~\citep{dubey2024llama3} as the base model in our experiments.
\paragraph{Synthetic Data} We use GPT-4o\footnote{gpt-4o-2024-08-06} to synthesize markup using the Asian Language Treebank (ALT) corpus \citep{AsianLanguageTreebank, 7918974}, generating 900 structured document pairs per language. ALT contains high-quality general domain parallel document units and aligns with our target language pairs.
The prompt includes a random example from the development set of SAP dataset and a sample of five tags from the target tag vocabulary.
Without domain-specific guidance, LLM defaults to generic tags (e.g., \texttt{<person>}), causing train-test mismatches. 
See full prompt in Appendix~\ref{sec:synth}.

\paragraph{Hyperparameters} The SFT model $\pi_\theta^{\text{SFT}}$ is then trained on both $100$ real and $0$ to $400$ synthetic structured document pairs using standard cross-entropy loss. We fine-tune for 20 epochs with batch size of $8$, a learning rate of $3 \times 10^{-7}$ and cosine learning rate scheduling with a warmup ratio of $0.1$. We use the AdamW optimizer~\cite{1711.05101}. Early stopping is triggered after $10$ evaluations without improvement, with evaluation performed every $10$ steps.

\subsubsection{Format Reinforcement}
We now describe the hyperparameter configuration for the reinforcement learning phase (\Section{sec:rl}), chosen based on our preliminary experiments.

\paragraph{Training Configuration.}
We report results using TreeSim reward in \Section{sec:main-res}, and results for Node-chrF and other rewards in \Section{sec:ablation-reward}.
We use a small learning rate of $10^{-6}$ and train for $5$ epochs with early stopping based on validation loss. Early stopping is triggered after $3$ evaluation steps without improvement, with evaluation and checkpointing performed every $3$ training steps. We set the maximum sequence length to $2,000$ tokens for both prompts and completions. The KL penalty coefficient $\beta$ is set to the default value of $0.01$. 
We select the checkpoint for testing based on the development set performance.

\paragraph{Batch and Generation Settings.}
We use $8$ generations per document ($K=8$) with a per-device batch size of $8$ and gradient accumulation steps of $1$, resulting in an effective batch size of $64$ across $8$ H200 GPUs. For generation, we use sampling with default temperature of $1.0$. 

\subsubsection{Computational Efficiency}
During training, we leverage DeepSpeed ZeRO-3 optimization and mixed precision training with bfloat16 for memory and computational efficiency. Each SFT model takes about $2.1$ hours and GRPO model takes about $1.3$ hours of training. We employ vLLM \citep{kwon2023efficient} for efficient inference where it takes $2$ minutes on test set.

\section{Results and Analysis}

\begin{table*}[t]
\centering
\resizebox{0.99\linewidth}{!}{%
\begin{tabular}{cccccccc@{\hspace{1em}}c}
\toprule
\multirow{2}{*}{\textbf{Src$\rightarrow$Tgt}} & \multirow{2}{*}{\textbf{Method}} & \multicolumn{2}{c}{\textbf{Translation}} & \multicolumn{2}{c}{\textbf{Structure}} & \multicolumn{2}{c}{\textbf{Combined}} & \cellcolor{gray!10}\multirow{2}{*}{\textbf{Avg.}} \\
\cmidrule(lr){3-4} \cmidrule(lr){5-6} \cmidrule(lr){7-8}
& & \textbf{Content-BLEU} & \textbf{Content-COMET} & \textbf{XML-Validity} & \textbf{XML-Match} & \textbf{XML-BLEU} & \textbf{StrucAUC} & \\
\midrule
\multirow{3}{*}{En$\rightarrow$Zh} & Prompt & 49.88 & 86.16 & 91.05 & 76.84 & 27.50 & 57.75 & 64.86 \\
                  & SFT & 49.66 & 86.47 & 94.21 & 85.26 & 36.38 & 63.57 & 69.26 \\
                  & \textbf{\formatrl{}} & \textbf{49.88} & \textbf{86.48} & \textbf{95.26} & \textbf{87.37} & \textbf{38.07} & \textbf{64.12} & \textbf{70.20} \\
\addlinespace[2.5pt]
\hdashline
\addlinespace[2.5pt]
\multirow{3}{*}{Zh$\rightarrow$En} & Prompt & 48.82 & 85.25 & 93.16 & 82.11 & 26.34 & 71.39 & 67.84 \\
                  & SFT & \textbf{56.41} & \textbf{85.34} & 94.74 & 83.68 & 27.58 & 71.66 & 69.90 \\
                  & \textbf{\formatrl{}} & 56.28 & 85.25 & \textbf{95.26} & \textbf{86.84} & \textbf{29.14} & \textbf{72.84} & \textbf{70.94} \\
\addlinespace[2.5pt]
\hdashline
\addlinespace[2.5pt]
\multirow{3}{*}{En$\rightarrow$Ja} & Prompt & 36.60 & 87.17 & 87.89 & 67.37 & 14.49 & 48.60 & 57.02 \\
                  & SFT & 39.11 & \textbf{88.22} & 95.26 & 84.21 & 27.47 & 60.40 & 65.78 \\
                  & \textbf{\formatrl{}} & \colorbox{sig05}{\textbf{39.30}} & 88.20 & \textbf{95.79} & \colorbox{sig05}{\textbf{88.42}} & \colorbox{sig05}{\textbf{30.32}} & \textbf{60.48} & \textbf{67.09} \\
\addlinespace[2.5pt]
\hdashline
\addlinespace[2.5pt]
\multirow{3}{*}{Ja$\rightarrow$En} & Prompt & 44.14 & 85.93 & 90.53 & 80.00 & 22.38 & 65.25 & 64.70 \\
                  & SFT & 52.19 & 85.96 & \textbf{95.26} & 82.11 & 24.15 & 67.92 & 67.93 \\
                  & \textbf{\formatrl{}} & \colorbox{sig05}{\textbf{52.79}} & \textbf{86.01} & 94.74 & \colorbox{sig05}{\textbf{87.37}} & \colorbox{sig05}{\textbf{26.67}} & \colorbox{sig05}{\textbf{69.82}} & \textbf{69.57} \\
\bottomrule
\end{tabular}%
}
\caption{Results of \formatrl{} and two baselines on structured documents. Bold indicates \textbf{the best performance}.
Background colors indicate statistical significance \footnotesize\colorbox{sig05}{$p<0.05$} compared to SFT.}
\label{tab:main-res}
\end{table*}

\begin{table*}[t]
\centering
\resizebox{0.99\linewidth}{!}{%
\begin{tabular}{cccccccc@{\hspace{1em}}c}
\toprule
\multirow{2}{*}{\textbf{Src$\rightarrow$Tgt}} & \multirow{2}{*}{\textbf{Method}} & \multicolumn{2}{c}{\textbf{Translation}} & \multicolumn{2}{c}{\textbf{Structure}} & \multicolumn{2}{c}{\textbf{Combined}} & \multirow{2}{*}{\textbf{Avg.}} \\
\cmidrule(lr){3-4} \cmidrule(lr){5-6} \cmidrule(lr){7-8}
& & \textbf{Content-BLEU} & \textbf{Content-COMET} & \textbf{XML-Validity} & \textbf{XML-Match} & \textbf{XML-BLEU} & \textbf{StrucAUC} & \\
\midrule
\multirow{3}{*}{En$\rightarrow$Zh} & Prompt & 54.79 & 85.87 & 96.32 & 84.74 & 43.33 & 56.62 & 70.28 \\
                  & SFT & \textbf{57.95} & 86.19 & \textbf{98.42} & 89.47 & 47.51 & 63.14 & 73.78 \\
                  & \textbf{\formatrl{}} & 57.70 & \textbf{86.22} & \textbf{98.42} & \textbf{90.00} & \textbf{47.63} & \textbf{64.29} & \textbf{74.04} \\
\addlinespace[2.5pt]
\hdashline
\addlinespace[2.5pt]
\multirow{3}{*}{Zh$\rightarrow$En} & Prompt & \textbf{39.92} & \textbf{83.31} & 95.26 & 83.68 & 33.83 & 67.06 & \textbf{67.18} \\
                  & SFT & 32.24 & 83.06 & 95.26 & 81.05 & 33.01 & 65.77 & 65.07 \\
                  & \textbf{\formatrl{}} & 34.76 & 82.83 & \textbf{95.79} & \colorbox{sig05}{\textbf{84.74}} & \textbf{34.74} & \textbf{66.28} & 66.52 \\
\addlinespace[2.5pt]
\hdashline
\addlinespace[2.5pt]
\multirow{3}{*}{En$\rightarrow$Ja} & Prompt & 40.13 & 87.90 & 96.32 & 79.47 & 26.78 & 45.07 & 62.61 \\
                  & SFT & 44.42 & 88.40 & 97.37 & 84.21 & 32.27 & 54.93 & 66.93 \\
                  & \textbf{\formatrl{}} & \colorbox{sig05}{\textbf{45.60}} & \textbf{88.60} & \textbf{98.42} & \textbf{86.84} & \colorbox{sig05}{\textbf{35.44}} & \textbf{55.04} & \textbf{68.32} \\
\addlinespace[2.5pt]
\hdashline
\addlinespace[2.5pt]
\multirow{3}{*}{Ja$\rightarrow$En} & Prompt & 35.61 & 84.86 & \textbf{98.95} & 81.58 & 27.58 & 64.65 & 65.54 \\
                  & SFT & 34.74 & 84.66 & 97.89 & 82.63 & 26.72 & 63.60 & 65.04 \\
                  & \textbf{\formatrl{}} & \colorbox{sig05}{\textbf{37.02}} & \textbf{84.96} & 98.42 & \colorbox{sig05}{\textbf{86.84}} & \colorbox{sig05}{\textbf{30.13}} & \colorbox{sig05}{\textbf{65.82}} & \textbf{67.20} \\
\bottomrule
\end{tabular}%
}
\caption{Results of \formatrl{} and two baselines on inline markup dataset. Bold indicates \textbf{the best performance}. Background colors indicate statistical significance \footnotesize\colorbox{sig05}{$p<0.05$} compared to SFT.}
\label{tab:inline-res}
\end{table*}

\subsection{Main Results}
\label{sec:main-res}
Table~\ref{tab:main-res} presents our main results on the structured document translation task across four language pairs. \formatrl{} using TreeSim reward consistently outperforms both the prompting and SFT baselines across nearly all evaluation metrics. Results of other rewards are shown in Appendix~\ref{sec:full-res-1}.

\paragraph{Structural Fidelity Improvements.}
\formatrl{} with TreeSim shows significant gains in structural preservation. XML-Match scores improve by an average of $3.69$ over SFT, with the largest improvement of $5.26$ points observed for Ja$\rightarrow$En. This indicates that \formatrl{} effectively learns to maintain document structure beyond what SFT achieves. 

\paragraph{Translation Quality Gains.}
Importantly, \formatrl{} maintains or slightly improves translation quality while enhancing structural fidelity. Content-\bleu{} scores increase by an average of $0.22$ points over SFT. Content-COMET scores remain stable, suggesting that our structural improvements do not come at the cost of translation quality.

\paragraph{Combined Performance.}
We show the combined improvement through XML-\bleu{}, which is widely used in previous work on structured data translation~\cite{salesforce-xml-bleu,dabre-etal-2024-effective}. It improves by $2.16$ points on average, and our proposed StrucAUC metric shows gains of $0.93$ points, confirming that improvements are robust across different structural error tolerances.

\paragraph{Human Evaluation.}
We performed a small-scale human evaluation on 60 rendered En$\rightarrow$Ja pages comparing {\formatrl{}} and prompting methods. For each page, an annotator compared the outputs of two methods against the reference, where the order of two outputs are randomly shuffled each time. Results show \formatrl{} won 29, prompting won 13, and 18 were ties. Qualitatively, outputs with (i) correct structure and (ii) correct embedded UI were preferred, suggesting that structural fidelity may be important in user experience.

\subsection{Results on Documents with Inline Markup}
\label{sec:inline-res}
Table~\ref{tab:inline-res} presents results on the inline markup dataset, where structural complexity is reduced to inline markup.
We found that although \formatrl{} with TreeSim still shows improvements in all metrics, the performance gap between the baseline method and \formatrl{} narrows considerably compared to structured documents. For example, the XML-Match gap between Prompt and \formatrl{} narrows from $10.92$ to $4.74$. This suggests that LLMs handle simpler inline structures effectively through in-context learning, but struggle with more structured documents. We show results of different rewards in Appendix~\ref{sec:full-res-2}.
\begin{figure}[t]
  \centering
  \includegraphics[width=0.98\columnwidth]{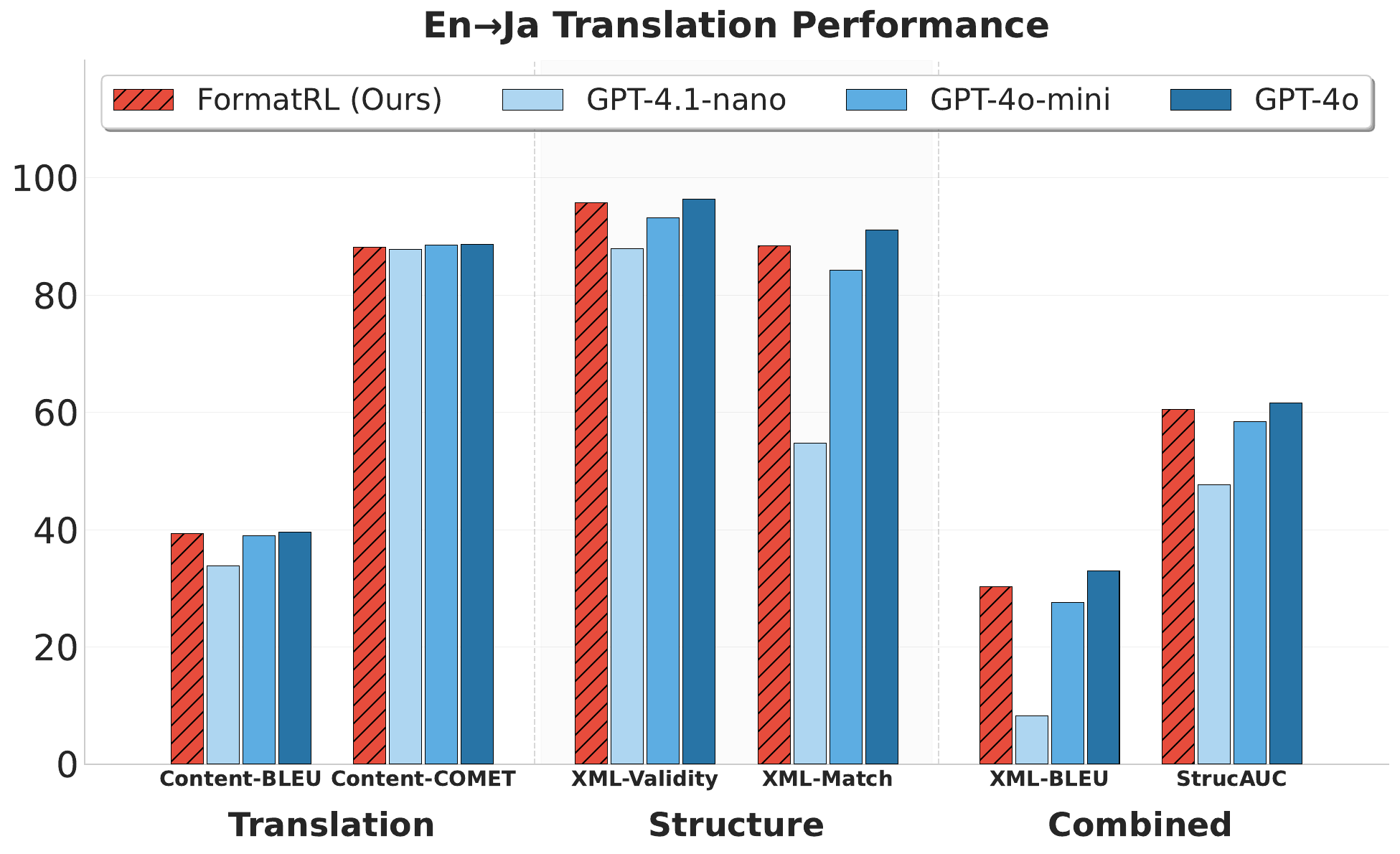}
  \caption{Comparison with GPT-4.1-nano {\small (2025-04-14)}, GPT-4o-mini {\small (2024-07-18)},  and GPT-4o {\small(2024-08-06)}.}
  \label{fig:gpt}
\end{figure}

\subsection{Comparison with GPT-4 Models}
We compare our approach to three GPT models which serve as reference. We show results of En$\rightarrow$Ja in ~\Figure{fig:gpt} and all directions in Appendix~\ref{sec:app_gpt}.
We found \formatrl{} shows comparable performance on most metrics with GPT-4o and outperforms GPT-4.1-nano and GPT-4o-mini.
Although with similar scores in automatic evaluation, after analyzing $60$ outputs of GPT-4o and our model, we found {\formatrl} outputs match the style (e.g. word choice is more formal) in source documents better than prompting with GPT-4o.

\subsection{Comparison with Parse-and-Assemble}
\begin{figure}[t]
  \centering
  \includegraphics[width=1\columnwidth]{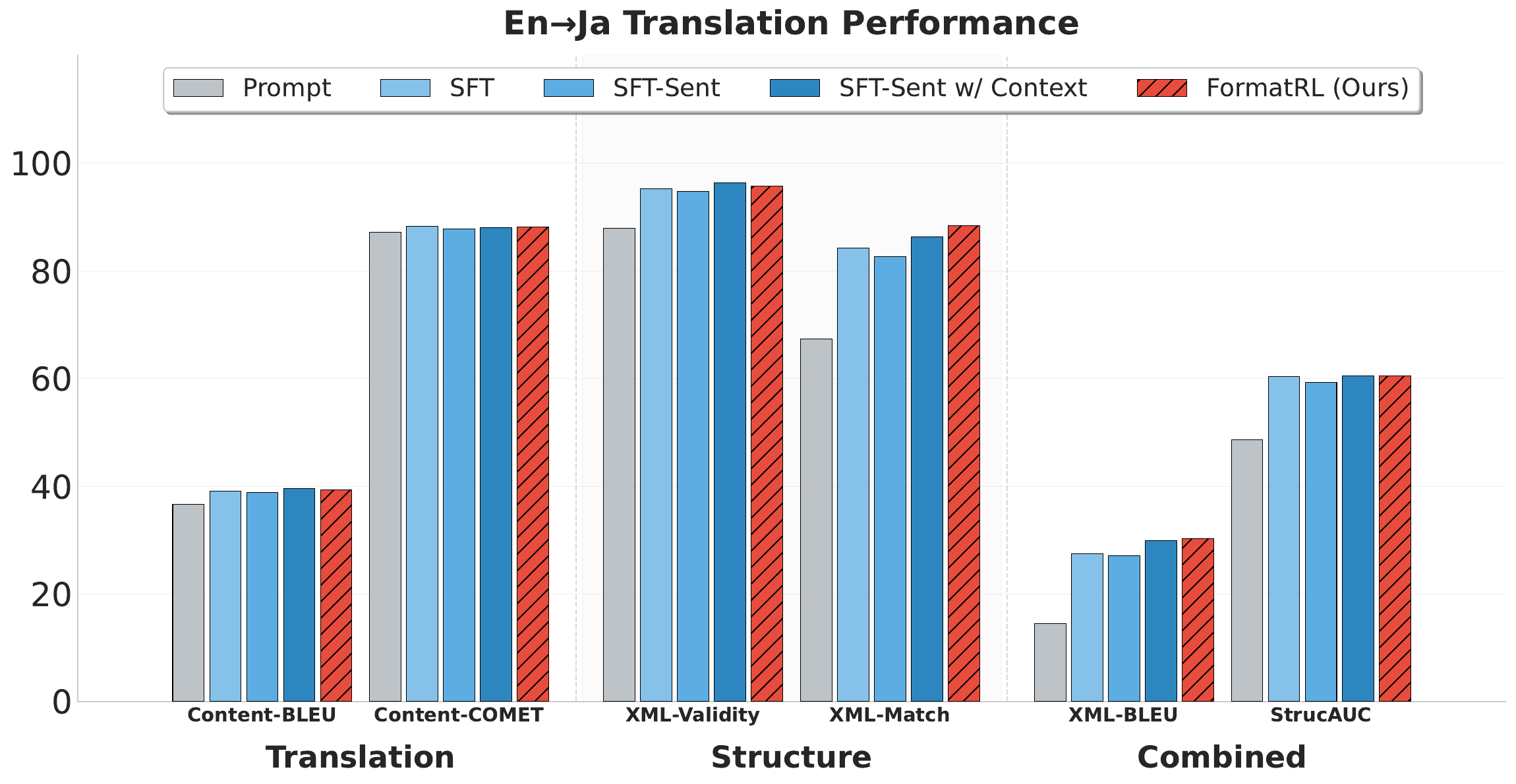}
  \caption{Comparison with parse-and-assemble baselines, in which the LLM acts as sentence-level MT model with or without document context.}
  \label{fig:baseline_sent}
\end{figure}

We implemented two parse-and-assemble baselines, where we first extract translatable text blocks, then apply an LLM-based sentence-level translator, and finally assemble the texts to form the output document.
SFT-Sent trains Llama 3.1 8B on parallel sentences whereas SFT-Sent w/ Content extents this by providing the whole document as context.
Figure~\ref{fig:baseline_sent} shows that for En→Ja, translation quality is comparable but \formatrl{} achieves higher XML-Match. 
Although parse-and-assemble ensures correct document structure, it struggles with in-line tags whose positions vary across target language syntax.
Furthermore, providing full documents for every sentence makes training 4.2$\times$ slower and inference 5.7$\times$ slower than standard SFT, showing that the end-to-end paradigm offers a more natural and efficient solution.

\begin{figure}[t]
  \centering
  \includegraphics[width=1\columnwidth]{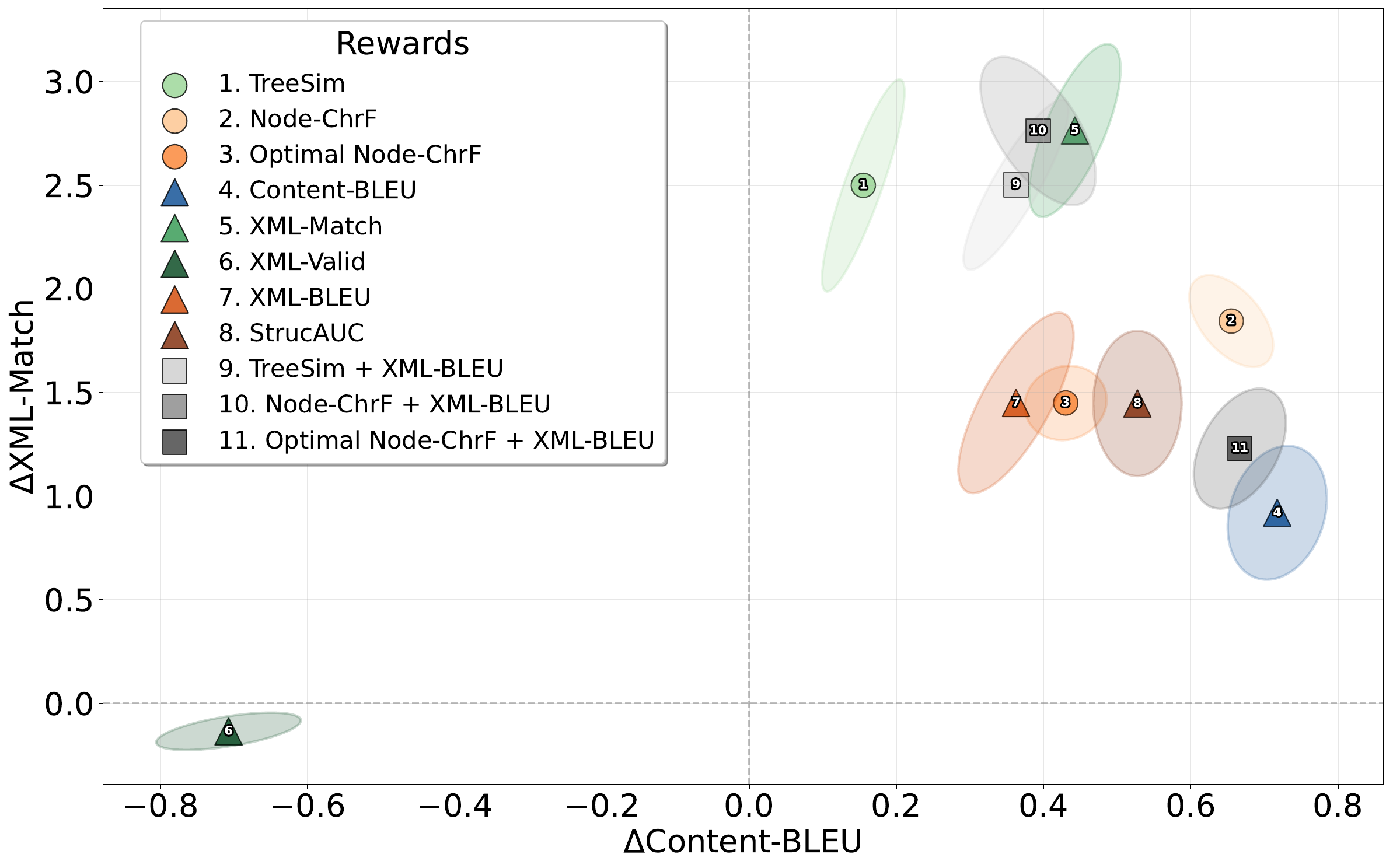}
  \caption{Improvement of \formatrl{} over SFT using various single rewards, and combinations of two rewards. Points represent mean improvement and ellipses visualize the local covariance directional structure between two metrics improvements.}
  \label{fig:ablation_reward}
\end{figure}

\subsection{Analysis: Reward Choice}
\label{sec:ablation-reward}
\Figure{fig:ablation_reward} shows the effect of different reward functions during GRPO training, including: 1) proposed TreeSim and Node-chrF, 2) metrics used in evaluation as rewards,\footnote{Content-\bleu and XML-\bleu here are document-level.} and 3) combination of two rewards. 
Estimates are constructed from $8$ runs of RL results. Refer to Appendix~\ref{sec:ablation-reward-appendix} for results featuring COMET instead of BLEU.

First, we found all rewards except XML-Validity to improve translation quality measured by Content-\bleu. Even pure structure-aware rewards, such as TreeSim and XML-Match, can improve translation. The combined reward Node-chrf improves both in good balance. However, not aligning to the reference (XML-Validity) is bad, hurting both translation and structure quality.
Second, the best way to optimize a specific metric is using it as a reward. Reinforcement learning with Content-\bleu as reward achieves the highest gain in Content-\bleu, and similarly, the XML-Match reward achieves the best XML-Match performance. Finally, we observe reward combination yields averaging effects, e.g., combining TreeSim with XML-BLEU shows better Content-\bleu improvement than TreeSim alone.

\subsection{Analysis: Reward-Metric Alignment}

Table~\ref{tab:treesim-nodechrf} compares the direct optimization effects of our two proposed rewards. We observe clear reward-metric alignment: using TreeSim as reward achieves the highest TreeSim scores, while Node-chrF reward yields the best Node-chrF scores in most directions. This confirms that reinforcement learning can effectively improve the specific structural properties defined by the reward.

\begin{table}[t]
\centering
\resizebox{\linewidth}{!}{%
\begin{tabular}{cccc}
\toprule
\textbf{Src$\rightarrow$Tgt} & \textbf{Method} & \textbf{TreeSim} & \textbf{Node-chrF} \\
\midrule
\multirow{5}{*}{En$\rightarrow$Zh}
& Prompt                 & 95.97 & 56.33 \\
& SFT                    & 97.86 & 62.67 \\
& \multicolumn{3}{l}{\formatrl{}} \\
& \quad \textit{w/ TreeSim}       & \textbf{98.19} & 63.24 \\
& \quad \textit{w/ Node-chrF}     & 97.70 & \textbf{64.41} \\
\addlinespace[2.5pt]
\hdashline
\addlinespace[2.5pt]
\multirow{5}{*}{Zh$\rightarrow$En}
& Prompt                 & 97.24 & 69.97 \\
& SFT                    & 97.54 & 70.58 \\
& \multicolumn{3}{l}{\formatrl{}} \\
& \quad \textit{w/ TreeSim}       & \textbf{98.12} & 71.77 \\
& \quad \textit{w/ Node-chrF}     & 97.97 & \textbf{73.00} \\
\addlinespace[2.5pt]
\hdashline
\addlinespace[2.5pt]
\multirow{5}{*}{En$\rightarrow$Ja}
& Prompt                 & 94.40 & 47.64 \\
& SFT                    & 97.55 & \textbf{59.77} \\
& \multicolumn{3}{l}{\formatrl{}} \\
& \quad \textit{w/ TreeSim}       & \textbf{97.98} & \textbf{59.77} \\
& \quad \textit{w/ Node-chrF}     & 97.23 & 59.26 \\
\addlinespace[2.5pt]
\hdashline
\addlinespace[2.5pt]
\multirow{5}{*}{Ja$\rightarrow$En}
& Prompt                 & 96.69 & 63.93 \\
& SFT                    & 97.82 & 66.47 \\
& \multicolumn{3}{l}{\formatrl{}} \\
& \quad \textit{w/ TreeSim}       & \textbf{98.28} & 68.51 \\
& \quad \textit{w/ Node-chrF}     & 98.02 & \textbf{68.93} \\
\bottomrule
\end{tabular}%
}
\caption{Performance comparison when optimizing TreeSim and Node-chrF rewards. Bold indicates the best performance.}
\label{tab:treesim-nodechrf}
\end{table}

\begin{figure}[t]
  \centering
  \includegraphics[width=1\columnwidth]{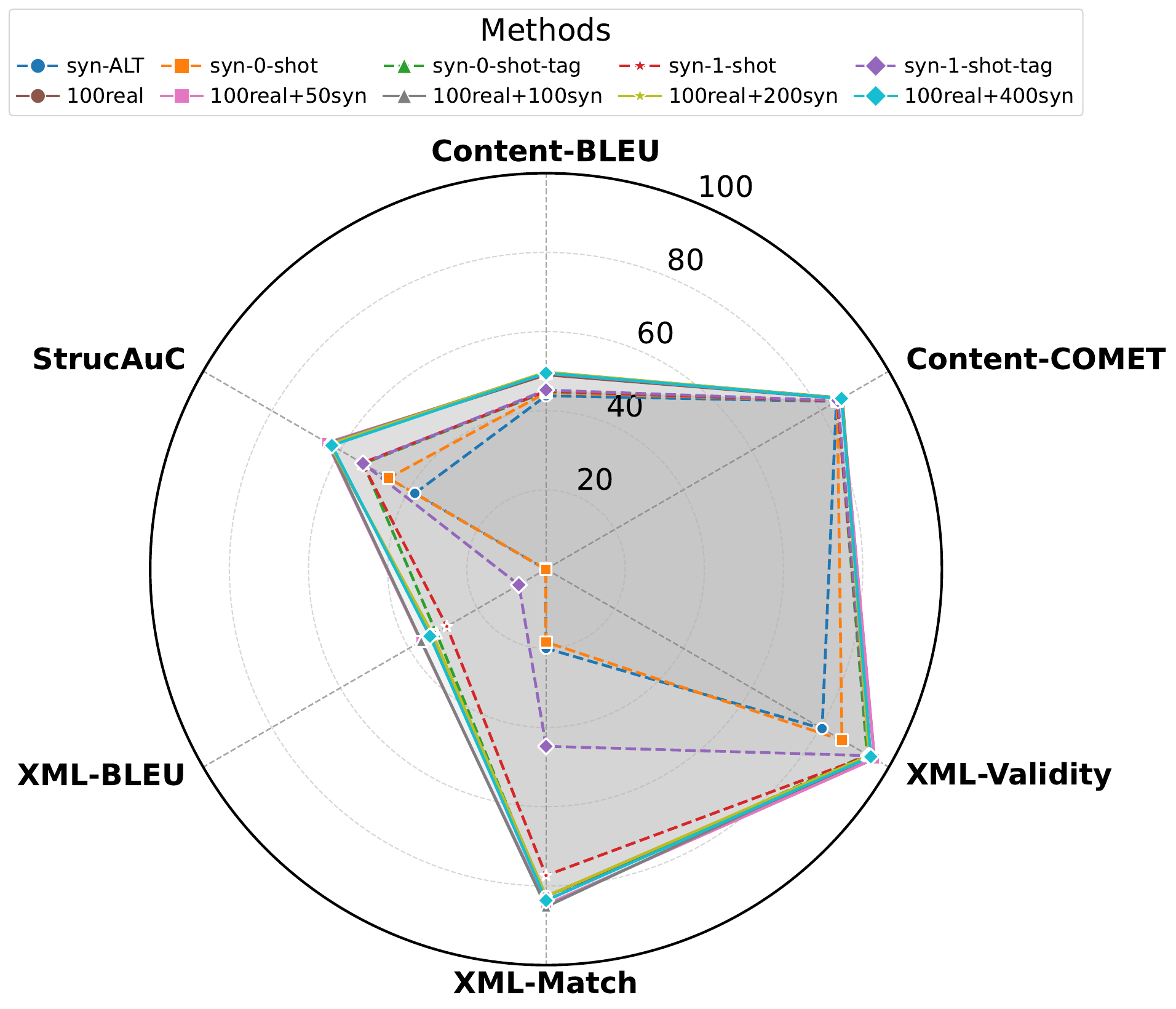}
  \caption{SFT Model performance comparison for English to Chinese translation by composition of training data. \textbf{syn-ALT} refers to fine-tuning using the raw ALT document pairs. \textbf{syn-0-shot} refers to data synthesized in a zero-shot manner. For \textbf{syn-1-shot}, the synthesizing LLM was provided with one example. The \textbf{*-tag} setups additionally guided the LLM with example XML tags from the development set. The \textbf{Xreal+Ysyn} setups are a mixture of real data and synthetic data generated with the syn-1-shot-tag approach.}
  \label{fig:radar_enja}
\end{figure}

\subsection{Analysis: Synthetic Data Strategies}
We explore the effect of using different synthetic data strategies to train the SFT model. 
As shown in \Figure{fig:radar_enja}, although the translation quality comes close to training on real data, using synthetic data alone can lead to catastrophic structure failure, with XML-Match scores dropping below $20\%$. 
We suppose that the domain shift in textual content likely has adverse interaction effects with the structural performance. Because it is unlikely the model completely independently learns structural transfer and translation.
This phenomenon highlights the crucial role of real target-domain XML markup. 
The importance of such markup is further underscored by observations that models trained on synthetic data generated without explicit guidance from in-domain examples and markup tags were more prone to structural errors. 
When combined with real data using the syn-1-shot-tag synthetic data, moderate amounts of synthetic data (e.g., 100real+100syn) can improve performance, whereas excessive amounts (e.g., 100real+400syn) can degrade it. For translation quality, this is not surprising: the Asian Language Treebank \citep{AsianLanguageTreebank} used for data generation differs substantially in domain from software documentation. 
Results of all language pairs are shown in \Figure{fig:radar_all_langs}.

\section{Conclusion}
To address the challenge of translating documents with complex structures, we propose \formatrl{}, a novel reinforcement learning approach with proposed structure-aware rewards: TreeSim and Node-chrF. We further propose StrucAUC as a fine-grained evaluation metric. Experimental results show \formatrl{} improves the structural fidelity of translated documents without compromising translation quality across both simple inline markup and complex structured documents.

\section{Limitations}

\paragraph{Limited Tag Set.} 
We restricted the tag set used during synthetic data generation to those present in the development set. While this approach provides consistency, it raises questions about the downstream translation models' ability to extrapolate to documents containing previously unseen tags. We did not evaluate this extrapolation capability due to budget constraints, as such an experiment would require generating substantially larger quantities of synthetic data with diverse markup using GPT. We did not explore tag abstraction using placeholder tags (e.g., \texttt{<t1>}) as which may help generalization but in the same time introduces pre-/post-editing and ignore semantics in human-interpretable tags.

\paragraph{Applying Sentence-level Metrics to Documents.} While we applied \bleu and COMET-22 to XML-stripped documents, these metrics, however, have known shortcomings when applied at the document-level as they are not designed/trained for such data \citep{jiang-etal-2022-blonde, vernikos-etal-2022-embarrassingly}.

\paragraph{Lack of Rigorous Human Evaluation.}
We performed a simple human evaluation in the result section. However, we are aware that a rigorous evaluation would require multiple annotators together with well-defined annotation instructions such as error taxonomies tailored to structured documents (e.g., MQM~\cite{freitag-etal-2021-experts} or ESA~\cite{kocmi-etal-2024-error} style annotation) that explicitly capture tag mismatch, nesting errors, and their severities. We leave this to future work.

\section*{Acknowledgements}
We would like to thank the reviewers for their insightful comments and suggestions. This work was supported by JSPS KAKENHI Grant-in-Aid for Early-Career Scientists 25K21290.
\bibliography{custom}

\clearpage
\appendix

\section{Synthetic Data Generation}
\label{sec:synth}
We show the prompt template used for synthetic data generation in \Figure{fig:syn_data_prompt}. It instructs GPT-4o to augment existing translation pairs without markup by inserting hierarchical XML markup elements into both source and target documents while maintaining alignment between the structures. In all $k$-shot settings we include full source–target document pairs as exemplars. 
For $k{=}5$, their combined length is about $\sim$9{,}800 characters ($\sim$5{,}070 Llama~3.1 tokens), which will differ for different language pairs.

\begin{figure*}[thb!]
\centering
\begin{tcolorbox}[title=Synthetic Data Generation Prompt, colback=gray!5!white, colframe=gray!75!black]
\small
Your task is to synthesize training data for machine translation of structured XML documents. Given a provided translation pair, insert well-aligned hierarchical XML markup into both source and target document. Do not translate the markup elements or include language codes. Here is an example of a well-aligned document pair:\newline\newline
SOURCE: \newline
<!DOCTYPE concept PUBLIC "-//SAP//DTD SAP DITA Composite//EN" "sap-ditabase.dtd">\newline
<concept id="loio16f62395d57f487e9937a092e4caefe9" xml:lang="en-US">\newline
<title>\newline
Download\newline
</title>\newline
<shortdesc>\newline
Downloads G/L account mappings into a .CSV file.\newline
...
\newline\newline
TARGET:\newline
<!DOCTYPE concept PUBLIC "-//SAP//DTD SAP DITA Composite//EN" "sap-ditabase.dtd">\newline
<concept id="loio16f62395d57f487e9937a092e4caefe9" xml:lang="en-US">\newline
<title>\newline
\begin{CJK}{UTF8}{min}ダウンロード\end{CJK}\newline
</title>\newline
<shortdesc>\newline
\begin{CJK}{UTF8}{min}G/L 勘定マッピングを.CSV ファイルにダウンロードします。
\end{CJK}file.\newline
...\newline
\newline
Now, insert markup into the following document pair. Output only the augmented source and target documents. Here are some example markup tags:\newline
<source></source>\newline
<uicontrol></uicontrol>\newline
<li></li>\newline
<p></p>\newline
<prolog></prolog>\newline
\newline
SOURCE:\newline
Every year around November 5th, people in Great Britain and some parts of the Commonwealth celebrate Guy Fawkes ...\newline
\newline
TARGET:\newline
\begin{CJK}{UTF8}{min}毎年11月5日前後に、グレートブリテンと連邦の一部地域の人々は、１６０５年１１月５日に国会議事堂を爆破することができなかったヨーク...\end{CJK}\newline
\end{tcolorbox}
\caption{Prompt template of \textbf{syn-1-shot-tag} used for data synthesis for the \textit{structured markup} translation task. This prompt features an example document pair from the development set as well as example tags sampled from development data to guide the LLM in data synthesis. For the \textbf{syn-1-shot} setup, the example tags are withhold. For the \textbf{syn-0-shot-tag}, the one-shot example is withhold. \textbf{syn-0-shot} features only the initial prompt and the data to synthesize from. For the \textit{inline} setup, the initial prompt is altered to 'Your task is to synthesize training data for machine translation of documents containing XML markup. Given a provided translation pair, insert well-aligned XML markup into both source and target document. Here is an example of a well-aligned document pair'.} 
\label{fig:syn_data_prompt}
\end{figure*}

\section{Example: Metrics Calculation}
\begin{figure*}[thb] 
  \centering
  \includegraphics[width=1\linewidth]{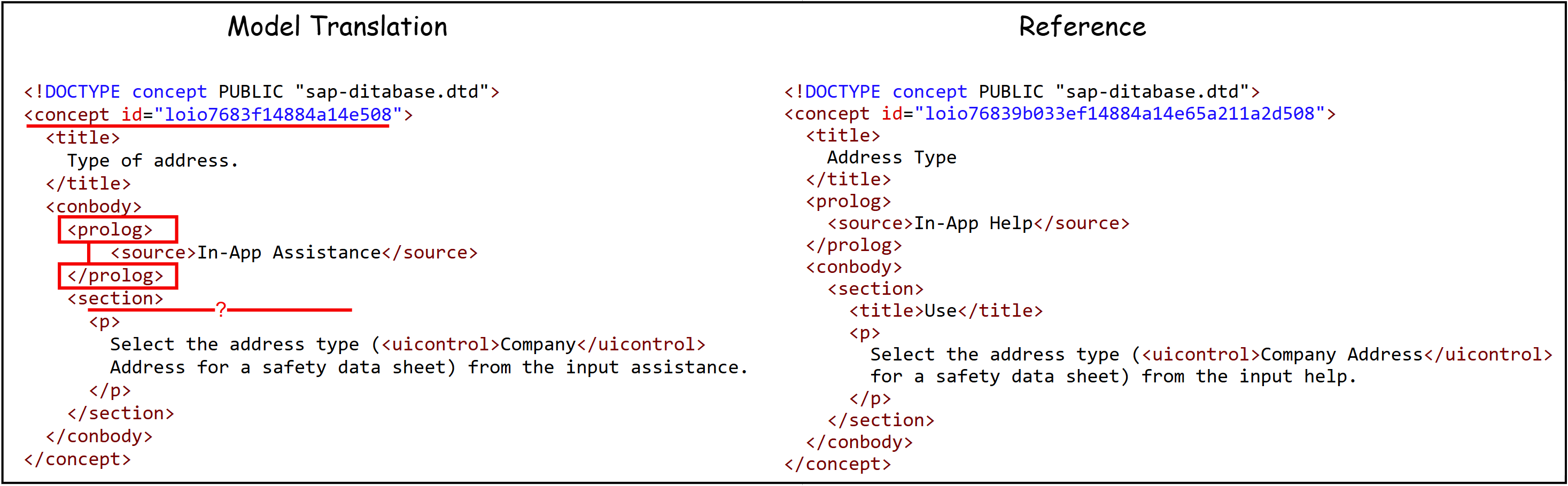}
  \caption{Toy-example of a model translation and reference with markings for purely structural errors.\newline 
  \textbf{XML-Validity}: The translation can be successfully parsed into an XML and therefore achieves a score of 1.\newline 
  \textbf{XML-Match}: The translation does not match the exact structure of the reference and therefore scores 0.\newline %
  \textbf{XML-\bleu}: Since the translation XML tree does not match the one of the reference the node contents of the reference will be paired with empty translations - e.g. (" ", "In-App-Help") - for corpus \bleu computation.\newline
  \textbf{StrucAUC}: The score is computed as a corpus level area under curve based on the respective \textit{Node-chrF} and \textit{Optimal Node-chrF}.\newline
  \textit{Node-chrF}: The structural errors of the translation will lead to misalignment in the parallel depth-first traversal. For instance, we will see a pairing of [...,(<conbody>,<prolog>), (<prolog>, <source>), (<source>, <conbody>)...] which overall results in a low node-level chrF score of 16.89.\newline
  \textit{Optimal Node-chrF}: With 3.5 edit operations (note that changing the label of the <concept> is considered half an edit), the nodes of the translation can be realigned to the reference, resulting in a Optimal Node-chrF of 52.92. \newline    
  }
  \label{fig:evaluation_example}
\end{figure*}

We illustrate how the metrics work using a toy example in~\Figure{fig:evaluation_example}, including XML-Validity, XML-Match, XML-\bleu, and the proposed StrucAUC (including the calculation of Node-chrF and Optimal Node-chrF).

\section{Full Results of Structured Document Setting}
\label{sec:full-res-1}
\Table{tab:structured-full} shows the detailed results of each reward function on the structured document setting across four language pairs (En$\rightarrow$ Zh, Zh$\rightarrow$ En, En$\rightarrow$ Ja, Ja$\rightarrow$ En). We found most of the rewards except XML-Validity improves across most metrics compared to supervised fine-tuning. Among the single reward functions, TreeSim usually achieves the best on structure score XML-Match, while Node-chrF shows the highest combined scores in three directions.

\section{Full Results of Inline Markup Setting}
\label{sec:full-res-2}
\Table{tab:inline-full} shows the detailed results of each reward function on the inline markup setting across four language pairs (En$\rightarrow$ Zh, Zh$\rightarrow$ En, En$\rightarrow$ Ja, Ja$\rightarrow$ En). 
We have similar observations that most of the rewards except XML-Validity improves across most metrics compared to supervised fine-tuning especially on structure and combined scores.

\section{Synthetic Data Performance: All Translation Directions}
\begin{figure*}[thb]
  \centering
  \begin{subfigure}[t]{0.48\textwidth}
    \centering
    \includegraphics[width=\linewidth]{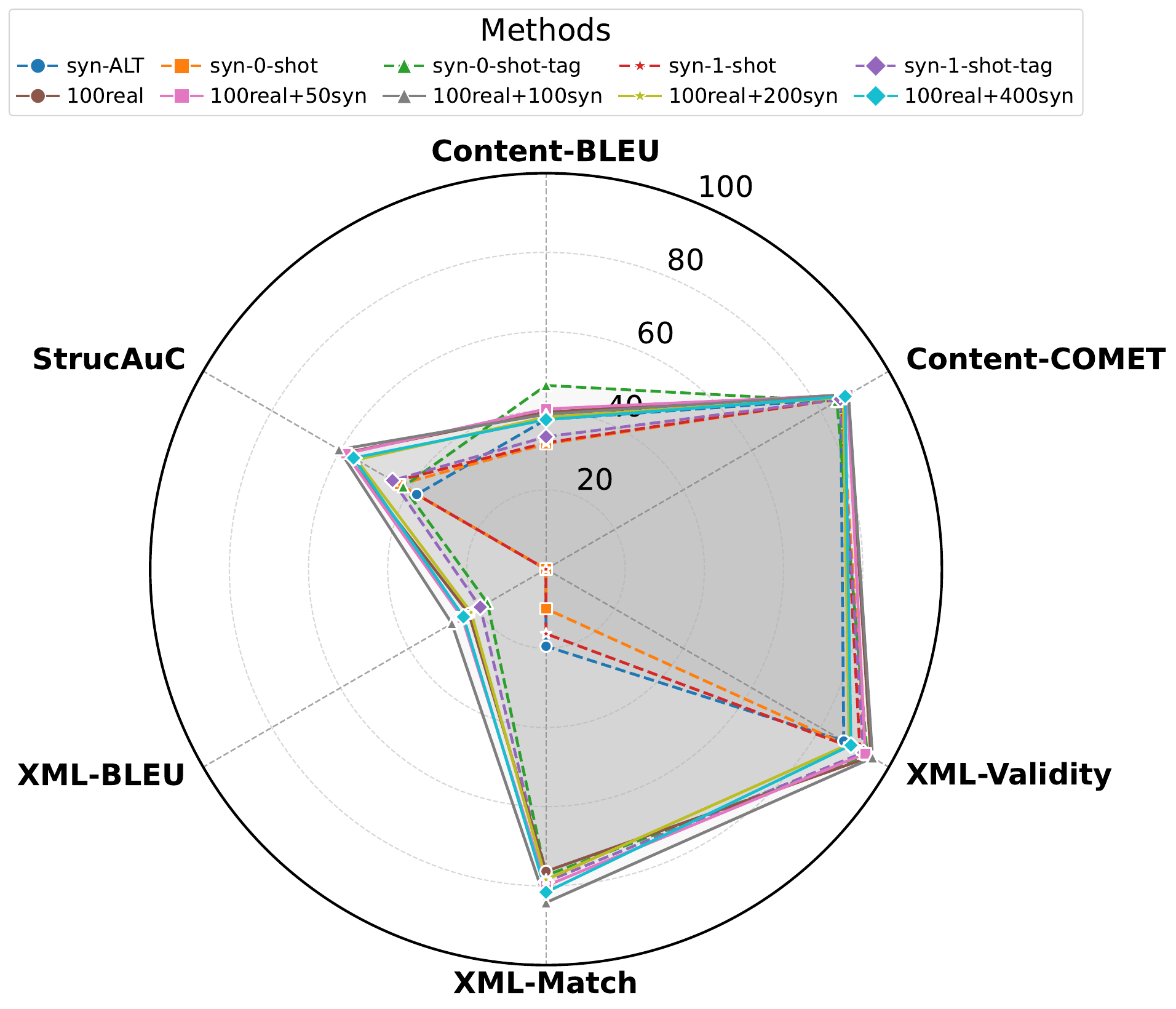}
    \caption*{\small \textbf{(a) enja}}
  \end{subfigure}
  \hfill
  \begin{subfigure}[t]{0.48\textwidth}
    \centering
    \includegraphics[width=\linewidth]{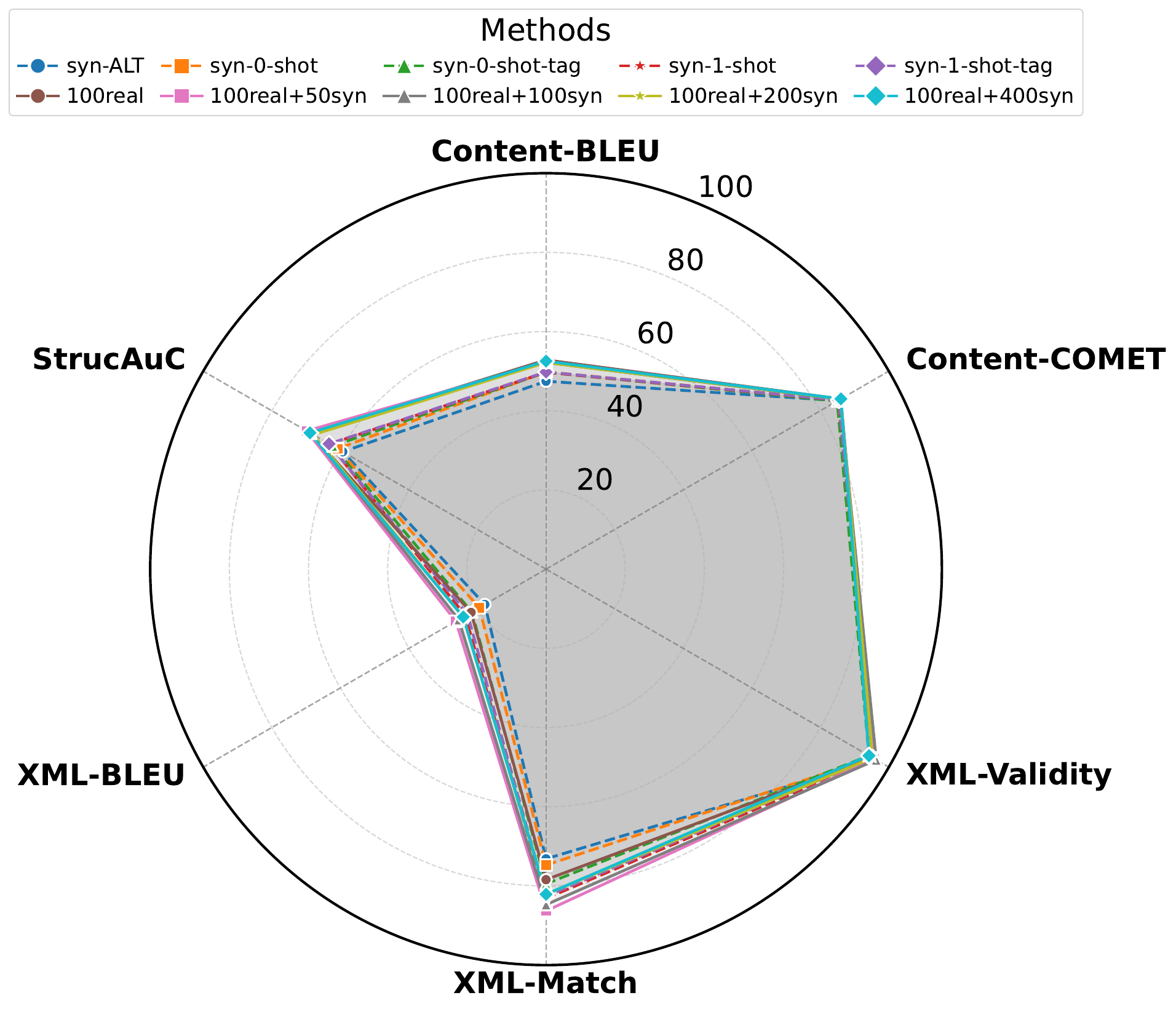}
    \caption*{\small \textbf{(b) jaen}}
  \end{subfigure}
  \vspace{0.5cm} 
  \begin{subfigure}[t]{0.48\textwidth}
    \centering
    \includegraphics[width=\linewidth]{fig/radar_chart_enzh.pdf}
    \caption*{\small \textbf{(c) enzh}}
  \end{subfigure}
  \hfill
  \begin{subfigure}[t]{0.48\textwidth}
    \centering
    \includegraphics[width=\linewidth]{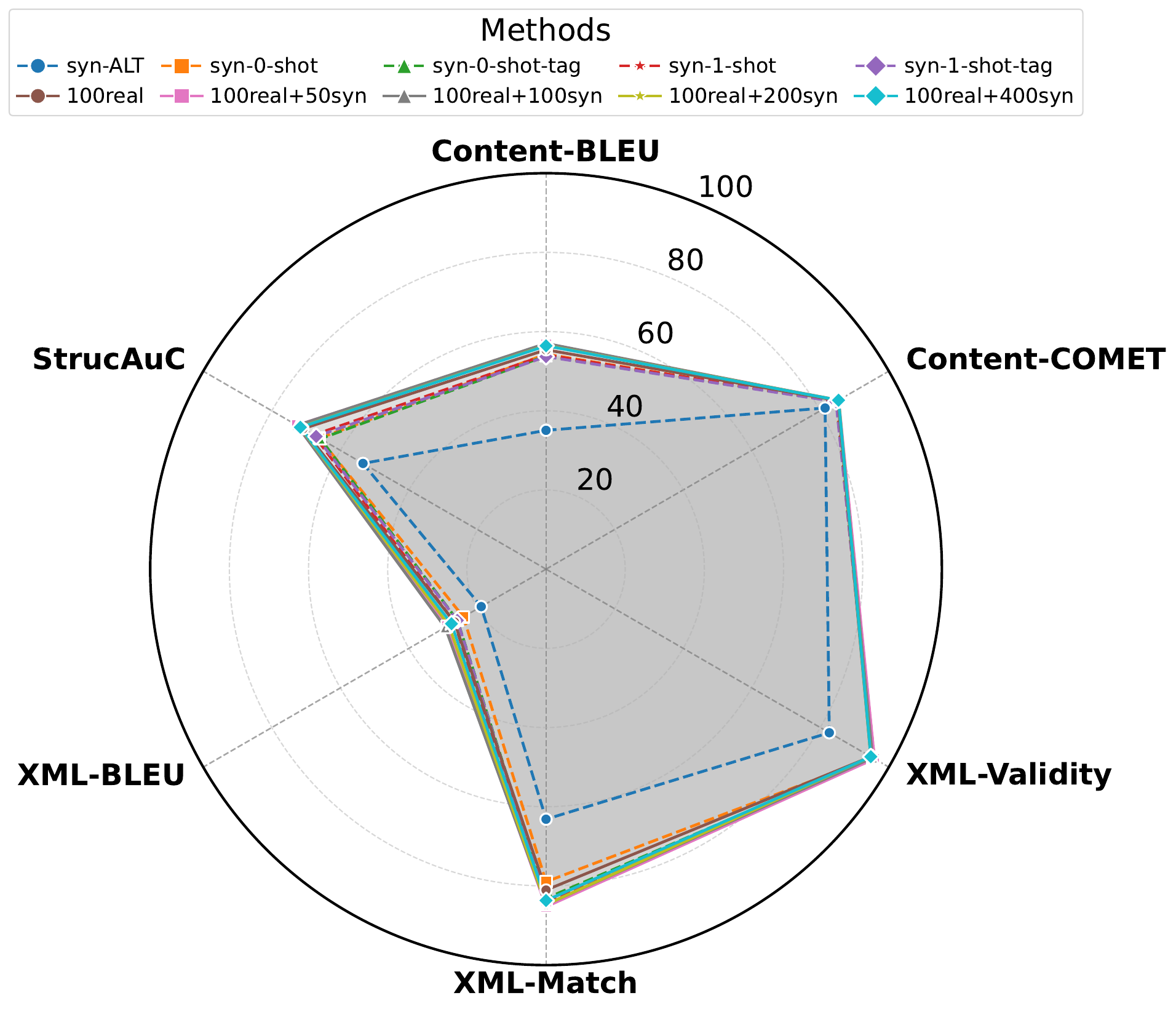}
    \caption*{\small \textbf{(d) zhen}}
  \end{subfigure}
  \caption{Comparison of performance of different synthetic generation methods used in supervised fine-tuning, across four language pairs.}
  \label{fig:radar_all_langs}
\end{figure*}

We show the effect of using different data when training the SFT model for all translation directions in~\Figure{fig:radar_all_langs}. 
The trends are similar: involving real data usually surpass pure synthetic data by a large margin (which is also expected). And the ratio of synthetic data and real data did not affect the performance that much. Usually 1:1 is a good balance, where too much synthetic data such as 4:1 slightly hurts the performance of the SFT model trained on such data.

\section{Comparison with GPT}
\label{sec:app_gpt}
\begin{figure*}[thb]
  \centering
  \setlength{\tabcolsep}{4pt}   %
  \begin{tabular}{cc}
    \includegraphics[width=0.48\linewidth]{fig/gpt_enja.pdf} &
    \includegraphics[width=0.48\linewidth]{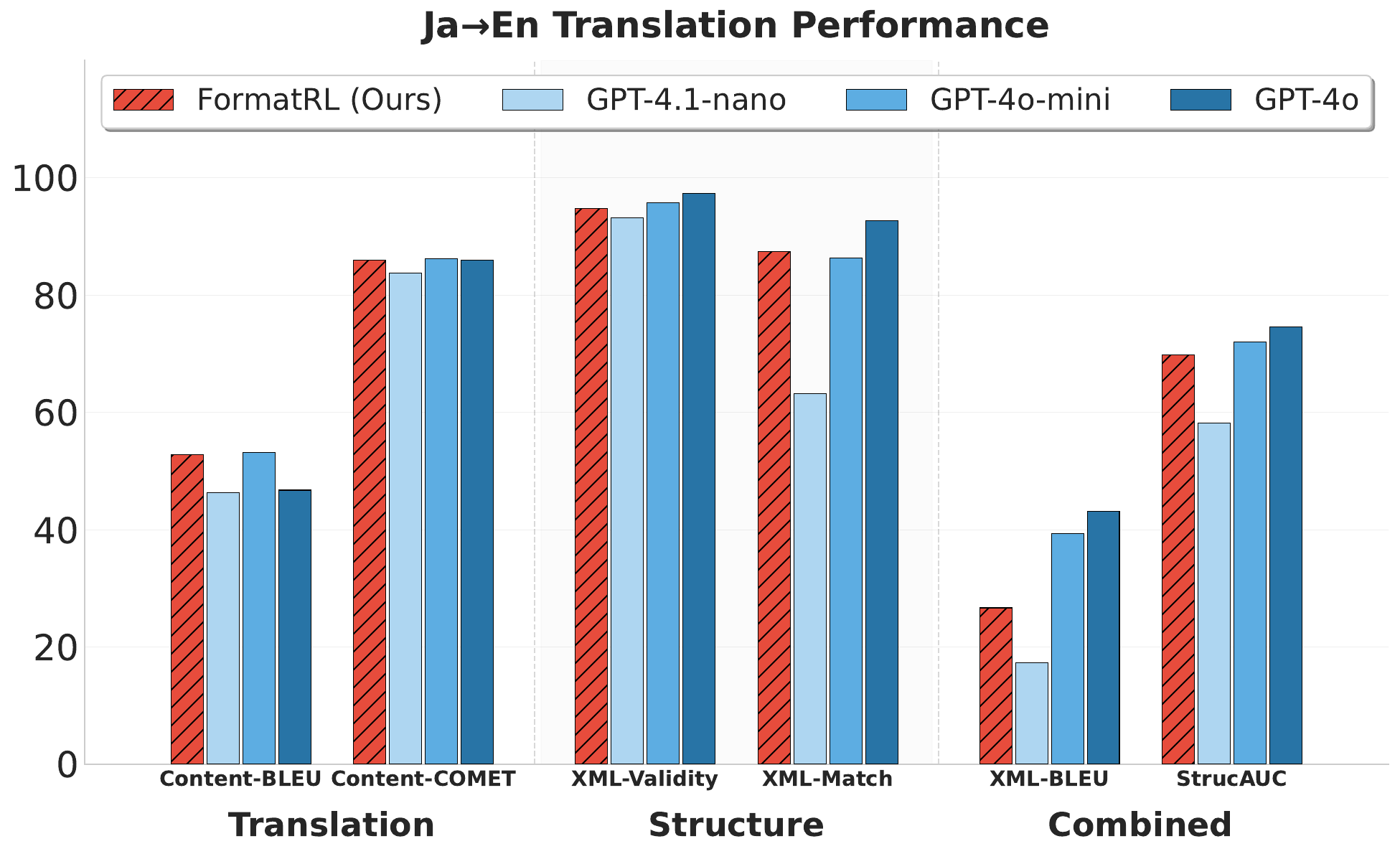} \\
    \includegraphics[width=0.48\linewidth]{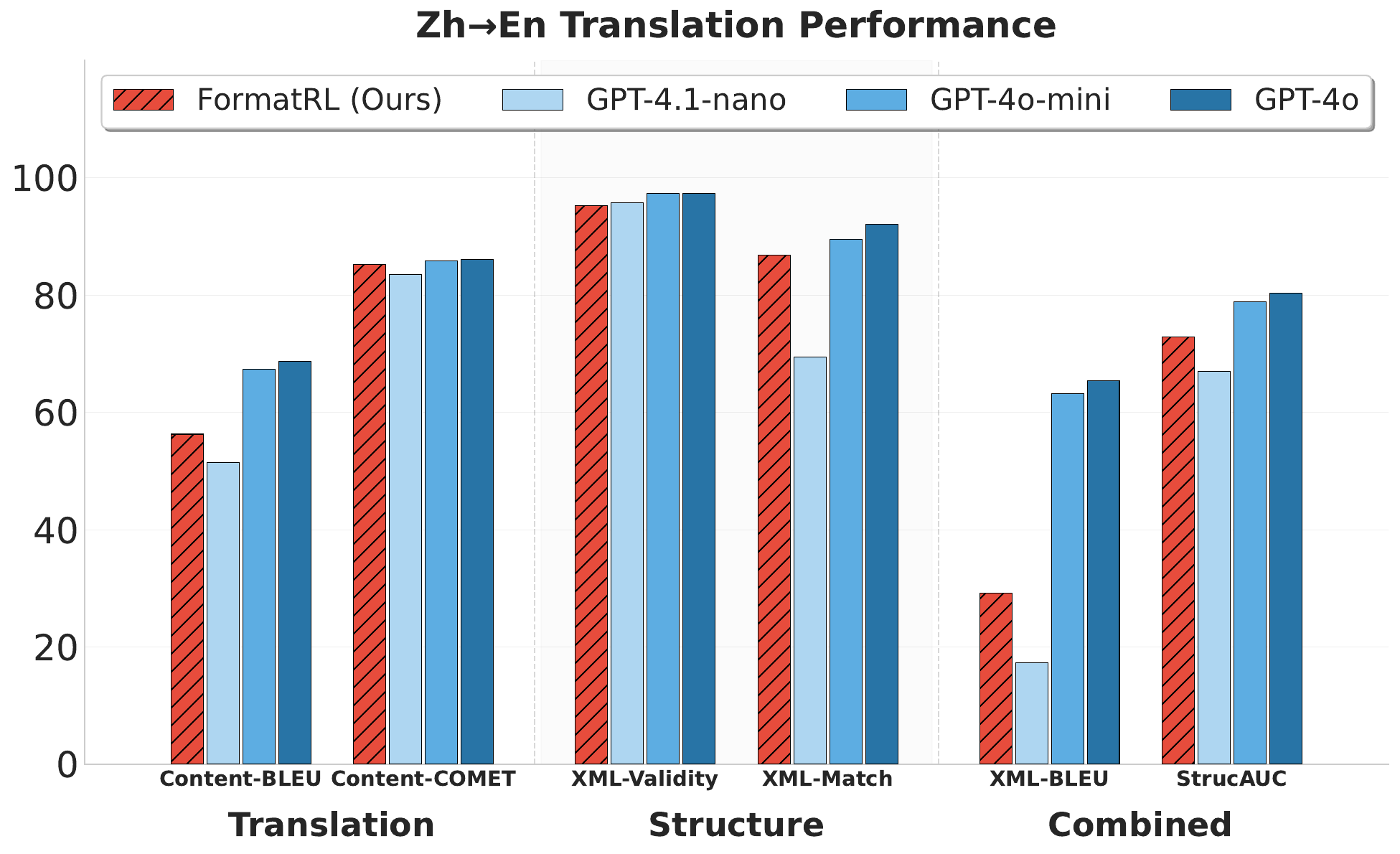} &
    \includegraphics[width=0.48\linewidth]{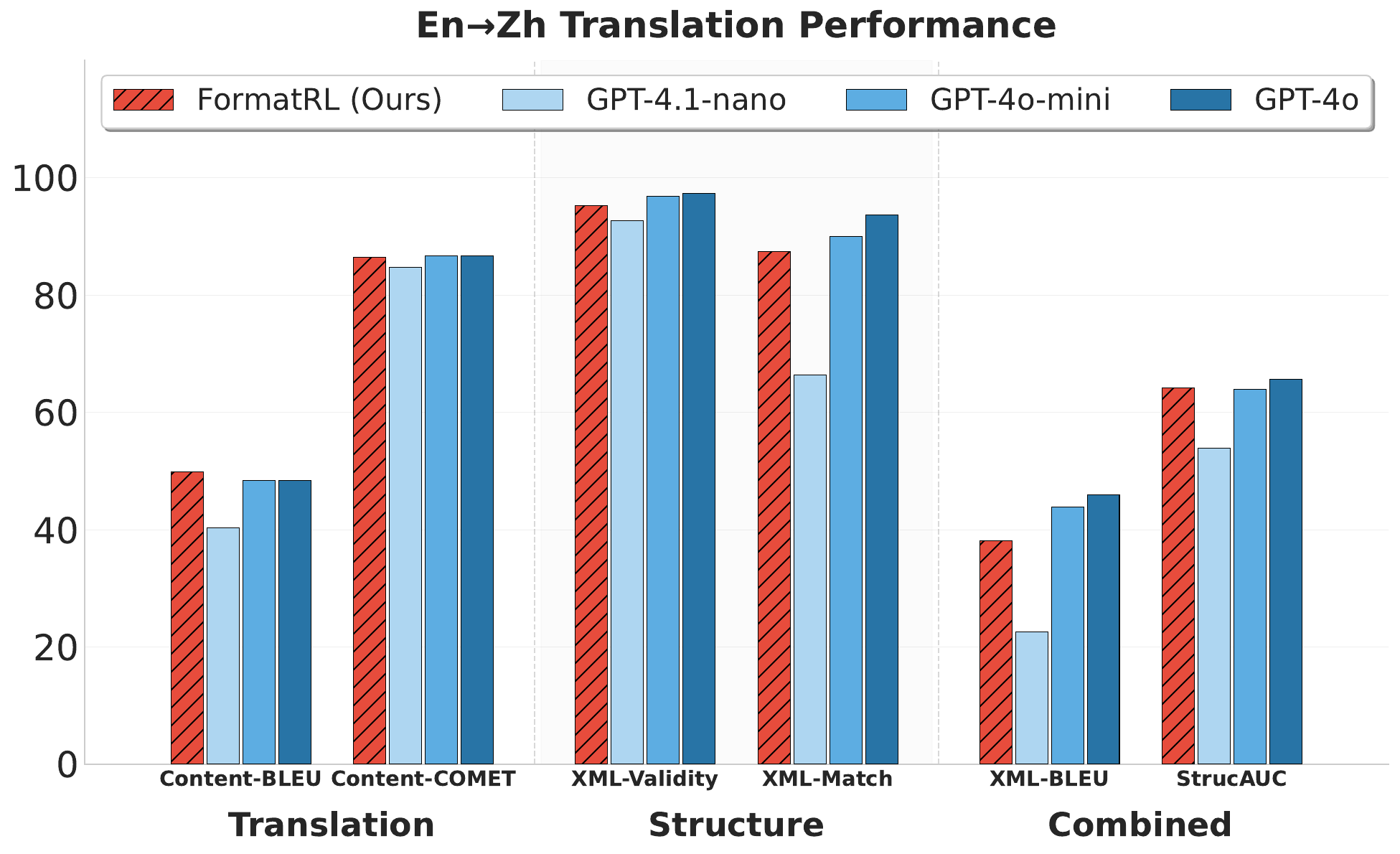} \\
  \end{tabular}
  \caption{Comparison with GPT models across four language pairs.}
  \label{fig:gpt_all}
\end{figure*}

We compare our approach to three GPT variants of different sizes in~\Figure{fig:gpt_all} which contains full results across all translation directions. \formatrl{} is usually better than GPT-4.1-nano, comparable to GPT-4o-mini, and not as good as GPT-4o. We found GPT-4o is especially strong at preserving XML markup, achieving the highest scores on structural and combined metrics.

\section{Ablation: Reward Choice on Content-COMET and XML-Match}\label{sec:ablation-reward-appendix}

\begin{figure}[thb]
  \centering
  \includegraphics[width=1\columnwidth]{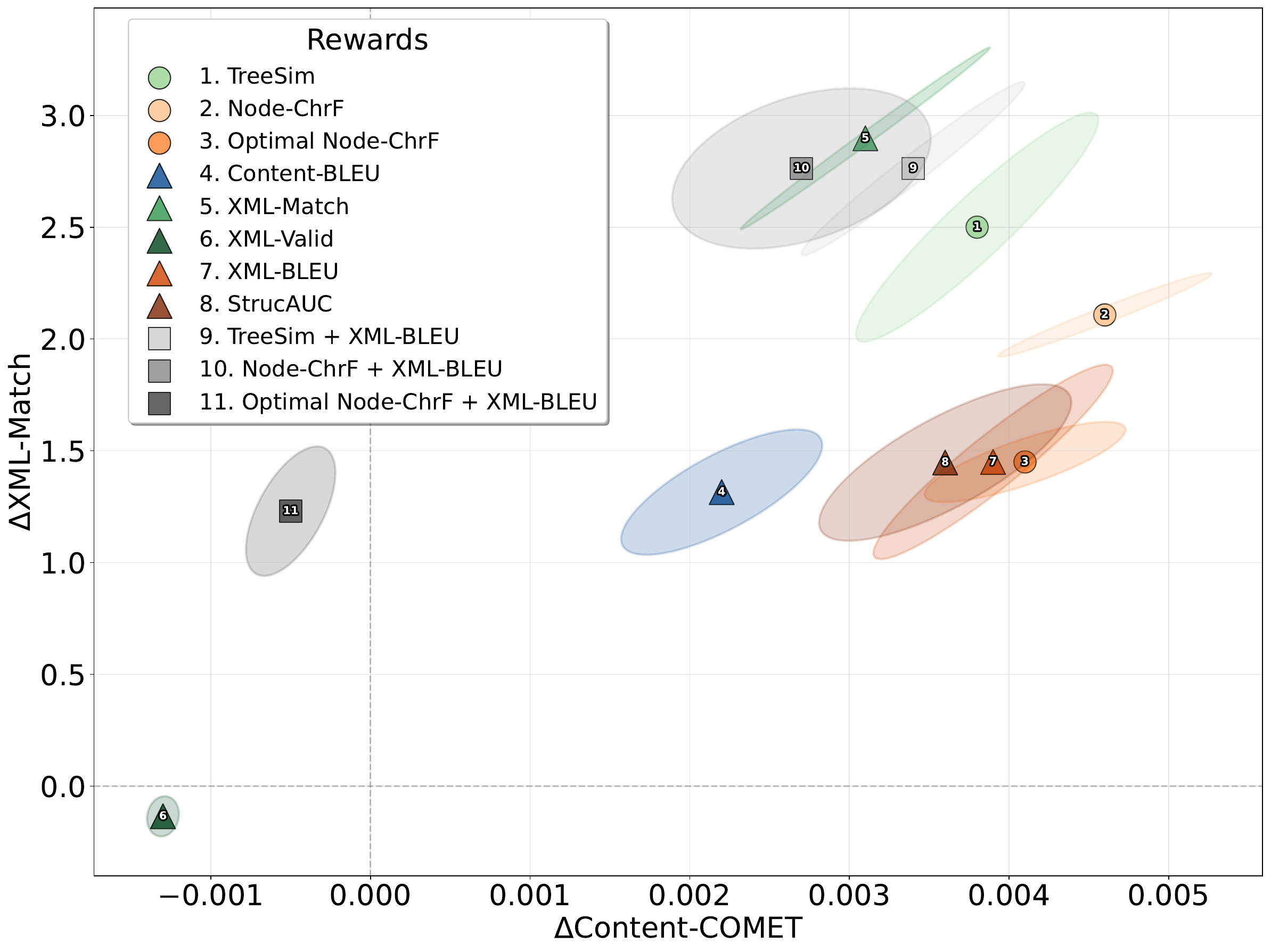}
  \caption{Improvement of \formatrl{} over SFT using various single rewards, and combinations of two rewards. Points represent mean improvement and ellipses visualize the local covariance directional structure between two metrics improvements. Estimates are constructed from RL results based on 8 SFT checkpoints each.}
  \label{fig:ablation_reward_2}
\end{figure}

Similar to \Figure{fig:ablation_reward}, \Figure{fig:ablation_reward_2} shows the effect of different reward functions during GRPO training using the improvement of Content-COMET instead of that of Content-BLEU.
We found this time using Content-BLEU as a reward function did not achieve the best improvement on Content-COMET, indicating the effect of reward overfitting: achieving the best on a given metric while it is promised to achieve the best on a similar metric (that measures the similar dimension).
Nevertheless, TreeSim achieved the highest XML-Match improvement among single rewards except using XML-Match itself, and Node-chrF achieved the highest Content-COMET improvement, indicating the effect of the proposed rewards.

\section{Details in StrucAUC}
\label{app:strucauc}
We show the StrucAUC algorithm in Algorithm~\ref{algo:strucauc}.
It is a very fast algorithm, the Hungarian matching is $O(n^3)$ in the number of text nodes $n$ per document, where in our dataset $n{<}20$, so matching costs are negligible versus model inference. Tree edit operations are linear in tree size under our restricted XML grammar.

\begin{algorithm*}[ht]
\caption{StrucAUC Metric}
\label{algo:strucauc}
\KwIn{Hypotheses $\{\hat{D}_{t,i}\}_{i=1}^n$, References $\{D_t^{\star,i}\}_{i=1}^n$, Maximum operations $K$}
\KwOut{StrucAUC score}

Initialize $S_k \gets \{\}$ for $k \in \{0, 0.5, 1, \ldots, K\}$\;

\For{$i = 1$ \KwTo $n$}{
    Parse $\hat{D}_{t,i}$ and $D_t^{\star,i}$ to XML trees\;
    \If{$D_t^{\star,i}$ invalid}{
        \textbf{continue}\;
    }
    \If{$\hat{D}_{t,i}$ invalid}{
        Add $0$ to all $S_k$ and \textbf{continue}\;
    }
    
    $s_{\text{unaligned}} \gets \text{Node-chrF}_{\text{parallel}}(\hat{D}_{t,i}, D_t^{\star,i})$\;
    $\mathcal{M}^* \gets \text{OptimalAlignment}(\hat{D}_{t,i}, D_t^{\star,i})$ \tcp{Hungarian algorithm}
    $d \gets \text{TreeEditDistance}(\hat{D}_{t,i}, D_t^{\star,i}, \mathcal{M}^*)$\;
    $s_{\text{optimal}} \gets \text{Node-chrF}_{\text{optimal}}(\mathcal{M}^*)$\;
    
    $S_0 \gets S_0 \cup \{s_{\text{unaligned}}\}$\;
    \For{$k \in \{0.5, 1, \ldots, K\}$}{
        \If{$d \leq k$}{
            $S_k \gets S_k \cup \{s_{\text{optimal}}\}$\;
        }
        \Else{
            $S_k \gets S_k \cup \{s_{\text{unaligned}}\}$\;
        }
    }
}

Compute AUC via trapezoidal integration over $\{(k/K, \text{mean}(S_k))\}$\;
\Return AUC $\times 100$\;
\end{algorithm*}

\section{Discussion}
We discuss some interesting aspects we think of our implementation for people who are interested in these details.

About the dataset, we wanted to try \formatrl{} on multiple datasets, after searching extensively for structured document translation datasets, we only found the SAP software document dataset. In the future we plan to curate some by ourselves. 

About hyper-parameters, we found GRPO does not require much training signal is the base SFT model has the basic structured document translation ability. In this case, the learning rate is a crucial parameter, we have tried learning rate from 1e-5 to 1e-7 and found 1e-6 is a good balance. Additionally, we save the checkpoint and evaluate it every 3 steps to capture the best one. Due to its efficiency, each training takes no more than $1.5$ hours and we in total spend less than $800$ GPU hours (100 hours in 8 H200 GPUs) for all GRPO experiments. For the memory efficiency, we found setting $K=8$, $Batch Size=8$, and max generation token of $800$ fits one H200 GPU with 141GB memory. 

In our analysis, we used Content-\bleu and XML-Match as reward, which may sounds like overfitting the metrics. However, the KL regularizer is added in the loss which prevents degenerate solutions that optimize only a single metric. Moreover, \textbf{our proposed novel metrics TreeSim and Node‑chrF do not overfit any metrics used in evaluation}.

\section{License}
We use the SAP software documentation dataset  which is under \textbf{The Creative Commons license
Attribution-Non Commercial 4.0 International (CC BY-NC 4.0)}, Asian Language Treebank (ALT) corpus under \textbf{The Creative Commons Attribution 4.0 International (CC BY 4.0) License}, and pre-trained models such as Llama-3.1-8B-Instruct under \textbf{The Llama 3.1 Community License} for research, which is consistent with their intended use. We have verified that the datasets do not contain personal information or offensive content.

We plan to release our code upon acceptance under \textbf{The Creative Commons license Attribution-Non Commercial 4.0 International (CC BY-NC 4.0)}. The code is intended for research purposes only and may not be used for commercial applications without explicit permission. 

\section{The Use of AI Assistants}
We used AI assistants for grammar and spelling checks. We sometimes also turn our incoherent listings of thoughts into a coherent paragraph which has always undergone further manual revisions.

\begin{table*}[tbh]
\centering
\resizebox{0.99\linewidth}{!}{%
\begin{tabular}{cccccccc}
\toprule
\multirow{2}{*}{\textbf{Src$\rightarrow$Tgt}} & \multirow{2}{*}{\textbf{Method}} & \multicolumn{2}{c}{\textbf{Translation}} & \multicolumn{2}{c}{\textbf{Structure}} & \multicolumn{2}{c}{\textbf{Combined}} \\
\cmidrule(lr){3-4}\cmidrule(lr){5-6}\cmidrule(lr){7-8}
& & \textbf{Content-BLEU} & \textbf{Content-COMET} & \textbf{XML-Validity} & \textbf{XML-Match} & \textbf{XML-BLEU} & \textbf{StrucAUC} \\
\midrule
\multirow{10}{*}{En$\rightarrow$Zh} & Prompt                & 49.88 & 86.16 & 91.05 & 76.84 & 27.50 & 57.75 \\
                  & SFT                   & 49.66 & 86.47 & 94.21 & 85.26 & 36.38 & 63.57 \\
& \multicolumn{7}{l}{\textbf{\formatrl{} \textit{w/ Reward of:}}} \\    
                  & TreeSim              & 49.88 & 86.48 & 95.26 & 87.37 & 38.07 & 64.12 \\
                  & Node-chrF            & 50.07 & 86.54 & 95.26 & 86.32 & 38.31 & 65.39 \\
                  & Node-chrF (opt.)      & 49.70 & 86.47 & 94.74 & 85.26 & 36.17 & 63.06 \\
                  & Content-BLEU         & 49.78 & 86.32 & 94.74 & 85.79 & 36.64 & 64.01 \\
                  & XML-Validity         & 49.31 & 86.21 & 95.26 & 85.26 & 36.17 & 63.65 \\
                  & XML-Match            & 49.75 & 86.37 & 95.79 & 85.26 & 35.97 & 64.20 \\
                  & XML-BLEU             & 49.38 & 86.41 & 95.26 & 84.74 & 35.56 & 63.29 \\
\addlinespace[2pt]\hdashline\addlinespace[2pt]
\multirow{10}{*}{Zh$\rightarrow$En} & Prompt                & 48.82 & 85.25 & 93.16 & 82.11 & 26.34 & 71.39 \\
                  & SFT                   & 56.41 & 85.34 & 94.74 & 83.68 & 27.58 & 71.66 \\
& \multicolumn{7}{l}{\textbf{\formatrl{} \textit{w/ Reward of:}}}   \\  
                  & TreeSim              & 56.28 & 85.25 & 95.26 & 86.84 & 29.14 & 72.84 \\
                  & Node-chrF            & 57.34 & 85.36 & 95.79 & 87.89 & 31.22 & 74.12 \\
                  & Node-chrF (opt.)      & 56.98 & 85.32 & 95.26 & 86.84 & 30.52 & 72.76 \\
                  & Content-BLEU         & 57.71 & 85.41 & 95.26 & 85.26 & 30.34 & 72.75 \\
                  & XML-Validity         & 55.94 & 85.16 & 94.74 & 86.32 & 28.73 & 71.42 \\
                  & XML-Match            & 56.39 & 85.16 & 94.74 & 87.89 & 30.31 & 72.33 \\
                  & XML-BLEU             & 57.35 & 85.49 & 94.74 & 87.37 & 30.17 & 73.56 \\
\addlinespace[2pt]\hdashline\addlinespace[2pt]
\multirow{10}{*}{En$\rightarrow$Ja} & Prompt                & 36.60 & 87.17 & 87.89 & 67.37 & 14.49 & 48.60 \\
                  & SFT                   & 39.11 & 88.22 & 95.26 & 84.21 & 27.47 & 60.40 \\
& \multicolumn{7}{l}{\textbf{\formatrl{} \textit{w/ Reward of:}}}   \\  
                  & TreeSim              & 39.30 & 88.20 & 95.79 & 88.42 & 30.32 & 60.48 \\
                  & Node-chrF            & 39.56 & 88.07 & 95.79 & 81.58 & 26.12 & 60.29 \\
                  & Node-chrF (opt.)      & 39.38 & 88.19 & 94.21 & 83.16 & 26.46 & 59.52 \\
                  & Content-BLEU         & 39.51 & 88.09 & 95.26 & 82.63 & 25.96 & 60.07 \\
                  & XML-Validity         & 39.12 & 88.07 & 95.26 & 83.68 & 26.05 & 59.63 \\
                  & XML-Match            & 39.39 & 88.15 & 95.26 & 87.37 & 29.56 & 60.36 \\
                  & XML-BLEU             & 39.71 & 88.19 & 94.21 & 86.32 & 28.22 & 60.15 \\
\addlinespace[2pt]\hdashline\addlinespace[2pt]
\multirow{10}{*}{Ja$\rightarrow$En} & Prompt                & 44.14 & 85.93 & 90.53 & 80.00 & 22.38 & 65.25 \\
                  & SFT                   & 52.19 & 85.96 & 95.26 & 82.11 & 24.15 & 67.92 \\
& \multicolumn{7}{l}{\textbf{\formatrl{} \textit{w/ Reward of:}}}                                                      \\ & TreeSim              & 52.79 & 86.01 & 94.74 & 87.37 & 26.67 & 69.82 \\
                  & Node-chrF            & 53.67 & 86.19 & 95.26 & 84.21 & 26.29 & 70.58 \\
                  & Node-chrF (opt.)      & 53.20 & 86.03 & 94.74 & 84.74 & 25.96 & 69.70 \\
                  & Content-BLEU         & 53.12 & 86.05 & 95.26 & 82.63 & 25.60 & 69.48 \\
                  & XML-Validity         & 52.43 & 85.94 & 95.26 & 82.63 & 24.37 & 69.05 \\
                  & XML-Match            & 53.12 & 85.98 & 95.26 & 86.32 & 26.45 & 69.26 \\
                  & XML-BLEU             & 53.53 & 86.07 & 94.21 & 85.26 & 26.67 & 69.79 \\
\bottomrule
\end{tabular}}
\caption{Full evaluation on structured documents, contrasting Prompt, SFT, and \formatrl{} with diverse reward functions.}
\label{tab:structured-full}
\end{table*}

\begin{table*}[thb]
\centering
\resizebox{0.99\linewidth}{!}{%
\begin{tabular}{cccccccc}
\toprule
\multirow{2}{*}{\textbf{Src$\rightarrow$Tgt}} & \multirow{2}{*}{\textbf{Method}} & \multicolumn{2}{c}{\textbf{Translation}} & \multicolumn{2}{c}{\textbf{Structure}} & \multicolumn{2}{c}{\textbf{Combined}} \\
\cmidrule(lr){3-4}\cmidrule(lr){5-6}\cmidrule(lr){7-8}
& & \textbf{Content-BLEU} & \textbf{Content-COMET} & \textbf{XML-Validity} & \textbf{XML-Match} & \textbf{XML-BLEU} & \textbf{StrucAUC} \\
\midrule
\multirow{9}{*}{En$\rightarrow$Zh} & Prompt              & 54.79 & 85.87 & 96.32 & 84.74 & 43.33 & 56.62 \\
                                   & SFT                 & 57.95 & 86.19 & 98.42 & 89.47 & 47.51 & 63.14 \\
                                   & \multicolumn{7}{l}{\textbf{\formatrl{} \textit{w/ Reward of:}}} \\
                                   & TreeSim             & 57.70 & 86.22 & 98.42 & 90.00 & 47.63 & 64.29 \\
                                   & Node-chrF           & 57.80 & 86.39 & 97.89 & 89.47 & 47.64 & 64.46 \\
                                   & Node-chrF (opt.)     & 56.95 & 86.10 & 98.42 & 88.42 & 45.28 & 63.13 \\
                                   & Content-BLEU        & 58.08 & 86.23 & 97.89 & 88.95 & 47.00 & 63.01 \\
                                   & XML-Validity        & 57.15 & 86.22 & 98.42 & 88.95 & 46.50 & 63.93 \\
                                   & XML-Match           & 57.86 & 86.26 & 98.42 & 88.42 & 46.80 & 65.45 \\
\addlinespace[2pt]\hdashline\addlinespace[2pt]
\multirow{10}{*}{Zh$\rightarrow$En} & Prompt              & 39.92 & 83.31 & 95.26 & 83.68 & 33.83 & 67.06 \\
                                   & SFT                 & 32.24 & 83.06 & 95.26 & 81.05 & 33.01 & 65.77 \\
                                   & \multicolumn{7}{l}{\textbf{\formatrl{} \textit{w/ Reward of:}}} \\
                                   & TreeSim             & 34.76 & 82.83 & 95.79 & 84.74 & 34.74 & 66.28 \\
                                   & Node-chrF           & 28.39 & 82.48 & 94.21 & 81.58 & 32.85 & 65.54 \\
                                   & Node-chrF (opt.)     & 29.71 & 82.60 & 94.21 & 80.53 & 32.50 & 65.11 \\
                                   & Content-BLEU        & 36.78 & 83.44 & 96.32 & 82.11 & 33.78 & 66.01 \\
                                   & XML-Validity        & 28.87 & 82.76 & 96.84 & 85.26 & 30.81 & 67.27 \\
                                   & XML-Match           & 35.87 & 82.65 & 94.74 & 80.53 & 32.03 & 65.94 \\
                                   & XML-BLEU            & 37.76 & 83.44 & 96.32 & 82.63 & 33.78 & 66.89 \\
\addlinespace[2pt]\hdashline\addlinespace[2pt]
\multirow{10}{*}{En$\rightarrow$Ja} & Prompt              & 40.13 & 87.90 & 96.32 & 79.47 & 26.78 & 45.07 \\
                                   & SFT                 & 44.42 & 88.40 & 97.37 & 84.21 & 32.27 & 54.93 \\
                                   & \multicolumn{7}{l}{\textbf{\formatrl{} \textit{w/ Reward of:}}} \\
                                   & TreeSim             & 45.60 & 88.60 & 98.42 & 86.84 & 35.44 & 55.04 \\
                                   & Node-chrF           & 45.04 & 88.50 & 97.37 & 86.84 & 34.90 & 54.89 \\
                                   & Node-chrF (opt.)     & 44.11 & 87.87 & 97.89 & 85.26 & 34.40 & 53.12 \\
                                   & Content-BLEU        & 45.72 & 88.62 & 98.42 & 85.79 & 34.86 & 56.08 \\
                                   & XML-Validity        & 44.44 & 88.45 & 98.42 & 86.32 & 34.26 & 54.12 \\
                                   & XML-Match           & 44.14 & 88.40 & 97.37 & 86.32 & 34.22 & 54.11 \\
                                   & XML-BLEU            & 45.53 & 88.58 & 98.42 & 87.37 & 36.97 & 54.29 \\
\addlinespace[2pt]\hdashline\addlinespace[2pt]
\multirow{10}{*}{Ja$\rightarrow$En} & Prompt              & 35.61 & 84.86 & 98.95 & 81.58 & 27.58 & 64.65 \\
                                   & SFT                 & 34.74 & 84.66 & 97.89 & 82.63 & 26.72 & 63.60 \\
                                   & \multicolumn{7}{l}{\textbf{\formatrl{} \textit{w/ Reward of:}}} \\
                                   & TreeSim             & 37.02 & 84.96 & 98.42 & 86.84 & 30.13 & 65.82 \\
                                   & Node-chrF           & 35.13 & 84.89 & 98.42 & 82.11 & 28.61 & 64.41 \\
                                   & Node-chrF (opt.)     & 32.61 & 84.91 & 98.95 & 83.16 & 28.98 & 64.46 \\
                                   & Content-BLEU        & 35.76 & 85.02 & 97.37 & 86.84 & 31.17 & 66.05 \\
                                   & XML-Validity        & 33.37 & 84.54 & 96.32 & 78.42 & 26.59 & 61.87 \\
                                   & XML-Match           & 33.50 & 84.72 & 97.89 & 82.63 & 27.66 & 64.09 \\
                                   & XML-BLEU            & 33.52 & 84.83 & 97.37 & 88.42 & 30.75 & 64.95 \\
\bottomrule
\end{tabular}}
\caption{Full evaluation on \textit{inline markup} documents. Each language pair lists Prompt, SFT, and \formatrl{} with different reward functions.}
\label{tab:inline-full}
\end{table*}

\end{document}